%% file: MNF.tex
\documentclass{article} 

\usepackage[nonatbib, preprint]{neurips_2025}
\usepackage[round]{natbib}

\usepackage[utf8]{inputenc} 
\usepackage[T1]{fontenc}    
\usepackage{hyperref}       
\usepackage{url}            
\usepackage{booktabs}       
\usepackage{amsfonts}       
\usepackage{nicefrac}       
\usepackage{microtype}      
\usepackage{xcolor}         
\usepackage{comment}

\usepackage{times}

\usepackage{import}

\input{contents/style.sty}
\input{contents/macros.sty}
\input{contents/shortcuts.sty}

 \author{%
 	Vijaya Krishna Yalavarthi* \\
 	ISMLL\\
 	University of Hildesheim\\
 	Germany \\
 	\texttt{yalavarthi@ismll.de} \\
 	\And
 	Randolf Scholz* \\
 	ISMLL\\
 	University of Hildesheim\\
 	Germany\\
 	\texttt{scholz@ismll.de}\\
 	\And
 	Christian Kloetergens \\
 	ISMLL\\
 	University of Hildesheim\\
 	Germany\\
 	\texttt{kloetergens@ismll.de}\\
 	\And
 	Kiran Madhusudhanan \\
 	ISMLL\\
 	University of Hildesheim\\
 	Germany\\
 	\texttt{madhusudhanan@ismll.de}\\
 	\And
 	Stefan Born\\
 	Institute of Mathematics\\
 	TU Berlin\\
 	Germany\\
 	\texttt{born@math.tu-berlin.de}
 	\And
 	Lars Schmidt-Thieme\\
 	ISMLL\\
 	University of Hildesheim\\
 	Germany\\
 	\texttt{schmidt-thieme@ismll.de}
}

\title{Marginalization Consistent Probabilistic Forecasting of Irregular Time Series via Mixture of Separable flows}
\begin{document}
\maketitle

\import{contents/}{abstract}

\import{contents/}{intro}
\import{contents/}{prelims}
\import{contents/}{related}
\import{contents/}{constructing}
\import{contents/}{prop_model}
\import{contents/}{experiments}
\import{contents/}{limitations}
\import{contents/}{conclusion}

\bibliographystyle{plainnat}
\bibliography{contents/neurips_24_mnf,contents/randolf}

\appendix
\import{contents/}{supplementary}

\newpage
\end{document}

%% file: contents/abstract.tex
\begin{abstract}
Probabilistic forecasting models for joint
distributions of targets in irregular time series with missing values are a heavily under-researched
area in machine learning with, to the best of our knowledge, only two
models are researched so far:
the Gaussian Process Regression model \citep{Durichen2015.Multitask},
and ProFITi \citep{Yalavarthi2024.Probabilistica}.
While ProFITi, thanks to using multivariate normalizing flows, is very expressive, leading to better predictive performance, it suffers from marginalization inconsistency:
it does not guarantee that the marginal distributions of a subset of variables in its predictive distributions coincide with the directly predicted distributions of these variables. When asked to directly predict marginal distributions, they are often vastly inaccurate.
We propose $\MyModel$ (Marginalization Consistent Mixture of Separable Flows),
a model that parametrizes a stochastic process through a mixture of several latent multivariate Gaussian Processes combined with separable univariate Normalizing Flows.
In particular, $\MyModel$ can be analytically marginalized allowing it to directly answer a wider range of probabilistic queries than most competitors.
Experiments on four datasets show that $\MyModel$ achieves both accurate joint and marginal predictions, surpassing all other marginalization consistent baselines, while only trailing slightly behind ProFITi in joint prediction, but vastly superior when predicting marginal distributions.
\end{abstract}

%% file: contents/intro.tex
\section{Introduction}\label{sec:intro}

In domains like whether and healthcare time series data is uneven: variables arrive at irregular intervals, channels are observed independently leading to extremely sparse time series when aligned.
While point prediction is the norm, many decision making applications require full probabilistic forecasts that capture uncertainty of possible outcomes.
To address this, researchers have developed probabilistic forecasting models for irregular time series \citep{DeBrouwer2019.GRUODEBayes, Deng2020.Modeling, Bilos2021.Neural, Schirmer2022.Modeling}. However, these models typically focus on univariate forecasts at single time points.

Yet many practical decisions ranging from diagnosing diseases to predicting weather depend on interactions between multiple variables over time, requiring accurate forecasts of joint multivariate distributions.
This area remains underexplored, with only two notable models: Gaussian Process Regression (GPR) \citep{Durichen2015.Multitask}, which models multivariate Gaussians, and ProFITi \citep{Yalavarthi2024.Probabilistica}, which uses normalizing flows for greater flexibility.
ProFITi achieves stronger performance but lacks a key property: marginalization consistency which guarantees that marginal distributions are the same whether queried directly or derived from the joint.

This consistency is crucial with varying numbers of observed variables. For instance, users ask a weather model for the probability of next three sunny days in San Diego and the chance of rain tomorrow.
If the answers contradict each other, trust in the model erodes—even if prediction of three sunny days is accurate.
In practice, we find that ProFITi, despite producing strong joint distributions, fails to maintain consistent marginals. On the other hand, GPR, while consistent, underperforms overall.

From this starting point we constructed a novel model that
combines the ideas of Gaussian Processes, normalizing
flows and mixture models in a way completely different from ProFITi and GPR,
to achieve both, guaranteed marginalization consistency
and high predictive accuracy (see Figure~\ref{fig:spaghetti}).
Overall our contributions as follows:
\begin{enumerate}

\item We formalize the underexplored property of marginalization consistency in probabilistic forecasting for irregular time series (Section~\ref{sec:prilims}). We propose Wasserstein Distance based metric to measure the marginalization inconsistency (Section~\ref{sec:mi_metric}).

\item We introduce a novel probabilistic forecasting model for irregular time series, \textbf{Marginalization Consistent Mixtures of Separable Flows} (\MyModel). \MyModel\ combines multiple normalizing flows with:
(i) \textbf{Gaussian Processes with full covariance matrices} as source distributions (as opposed to the usual identity matrix), and
(ii) a \textbf{separable} invertible transformation, applied independently per dimension rather than jointly.
We formally prove that $\MyModel$ is guaranteed to be Marginalization Consistent (Sections~\ref{sec:mc_layers} and~\ref{sec:mix_flows}).

\item In experiments on four datasets, we show that $\MyModel$ outperforms other state-of-the-art marginalization-consistent models in both multivariate joint and univariate marginal distributions. While its performance on joint distributions is comparable to or slightly below that of ProFITi, \MyModel\ significantly surpasses ProFITi in univariate marginals (Section~\ref{sec:exp}), demonstrating the advantage of Marginalization Consistency.
{Code available at~\url{https://anonymous.4open.science/r/seperable_flows-BACC}}
\end{enumerate}

%% file: contents/prelims.tex
\section{Preliminaries}\label{sec:prilims}

We use the triplet representation of an irregular time series $\X$, which is a sequence of $N$-many triplets~\citep{Horn2020.Set, Yalavarthi2024.Probabilistica}:
\begin{align}\label{eq:context_space}
	\begin{aligned}
	\X := \bigl( (t_n^\obs, c_n^\obs, v_n^\obs) \bigr)_{n=1:N}\in\Seq(\mathcal{X}), 
	\quad \mathcal{X} = \R\times\{1,\ldots,C\}\times\R
	\end{aligned}
\end{align}
where ${t_n^\obs \in \R}$ is the observation time point, and ${v_n^\obs \in \R}$ is the observed value in channel ${c_n^\obs \in \{1, \ldots, C\}}$.
A \textbf{time series query} $\Q$ is a sequence of $K$-many pairs:
\begin{align}\label{eq:query_space}
	\begin{aligned}
	\Q := \bigl((t_k^\qry, c_k^\qry)\bigr)_{k=1:K}\in\Seq(\mathcal{Q}), \quad \mathcal{Q} = \R\times\{1,\ldots,C\}
	\end{aligned}
\end{align}
where ${t_k^\qry}$ is the future time point and ${c_k^\qry}$
is the queried channel.
A \textbf{forecasting answer} $y$ is a sequence of scalars: ${y = (y_1, \ldots, y_K)}$, where $y_k \in \R$ is the forecasted value in channel ${c_k^\qry}$ at time ${t_k^\qry}$. Here, $\Seq(\mathcal{X})$ denotes the space of finite sequences over $\mathcal{X}$.
All the query time points are after the observations: ${\min_{k=1:K} t_k^\qry > \max_{n=1:N}t_n^\obs}$.

\paragraph{Requirements.} A marginalization consistent probabilistic irregularly sampled
time series forecasting model must satisfy the following requirements:
\begin{enumerate}[%
	label={\textbf{R\arabic*}},
	align=right,
	leftmargin=*,
]%
\item\label{req:joint_prediction} \textbf{Joint Multivariate Prediction}.
The task of probabilistic irregular time series forecasting is to find a model $\hat{p}$ that can predict the joint multivariate distribution $\hat{p}(y\mid \Q, \X)$ of the answers $y$, given the query points $\Q$ and observed series $\X$. Both the context $N = |X|$ and the query length $K = |Q|$ are allowed to be dynamic.
	\begin{align}\label{eq:forecasting_model}
	\begin{aligned}
		& \hat{p}\colon \Seq(\R\times\mathcal{Q})\times\Seq(\mathcal{X}) \to \R_{\ge 0},
		\\ (y, \Q, \X) \mapsto & \hat{p}(y_1, \ldots, y_K \mid Q_1, \ldots, Q_K, X_1, \ldots, X_N)
	\end{aligned}
	\end{align}
	So that, for a given pair $(\Q, \X)$, the partial function $(y_1, \ldots, y_K) \mapsto \hat{p}(y_1, \ldots, y_K \mid \Q, \X)$ realizes a probability density on $\R^{|Q|}$.
\item\label{req:permutation_invariance} \textbf{Permutation Invariance}.
	As the time stamp and channel-ID are included in each sample, the order of the samples does not matter, and hence any model prediction should be independent of the order of both the query or context:
	\begin{align}\label{eq:permutation_invariance}
		\hat{p}(y\mid \Q, X) = \hat{p}(y^\pi \mid \Q^\pi, \X^\tau),
		\, \forall\pi\in S_{|\Q|}, \tau\in S_{|\X|}
	\end{align}
\item\label{req:mar_consistency} \textbf{Marginalization Consistency/Projection Invariance}.
	Predicting the joint density for the sub-query $\Q_{-k}$ given by removing
	the $k$-th item from $\Q$ should yield the same result as marginalizing the $k$-th variable from the complete query $\Q$.
	\begin{align}\label{eq:mar_consistency}
		\hat{p}(y_{-k} \mid \Q_{-k}, \X)
		& = \int_{\R} \hat{p}(y \mid \Q, \X) \dd{y_k}
	\end{align}
	This generalizes to any subset $K_S\subseteq \{1,\ldots,K\}$.
\end{enumerate}
For a model satisfying~\ref{req:joint_prediction}-\ref{req:mar_consistency},
we will only have to marginalize if we try to validate the marginalization consistency.
For this validation we added requirement~\ref{req:mar_consistency}.
\citet{Yalavarthi2024.Probabilistica} discussed~\ref{req:joint_prediction} and~\ref{req:permutation_invariance}, but did not consider~\ref{req:mar_consistency}. We argue that irregularly sampled time series is realization of a stochastic process and~\ref{req:mar_consistency} is a fundamental property of any model that mimics it.
\begin{theorem}\label{thm:kolmogorov_extension}
Any model that satisfies~\ref{req:joint_prediction}-\ref{req:mar_consistency} realizes an $\R$-valued stochastic process over the index set $T=\R\times\{1,\ldots,C\}$. \\
Proof. This is a direct application of Kolmogorov's extension theorem~\citep{oksendal_stochastic_differential_equations_2003}
\end{theorem}
Marginalization consistency provides performance guarantees: when querying a consistent model, that is known to be close to the ground truth for queries of size $K$, then it also produces predictions close to the ground truth for queries of size $<K$. This is a consequence of the data processing inequality (DPI;~\citealp{murphy_probabilistic_machine_learning_2022}).
\begin{align*}
	&D_{KL}\Bigl(p(y_1, \ldots, y_K\mid Q_1,\ldots, Q_K,X)\mid\hat{p}(y_1, \ldots, y_K\mid Q_1, \ldots, Q_K, X)\Bigr)
	\\ & \geq D_{KL}\Bigl(p(y_1, \ldots, y_{K-1}\mid Q_1, \ldots, Q_{K-1}, X)\mid \hat{p}(y_1, \ldots, y_{K-1}\mid Q_1, \ldots, Q_{K-1},X)\Bigr) \quad (\ref{req:mar_consistency})
\end{align*}
Hence, we expect marginalization consistent models to generalize better across query sizes.
This is directly reflected by the experimental results in Table~\ref{tab:irreg_forec} and~\ref{tab:mnll}: models, that are not consistent, do not perform well on the marginal prediction task.

%% file: contents/related.tex
\section{Related Work}\label{sec:related}

There have been multiple works that deal with point forecasting of irregular time series~\citep{Ansari2023.Neural,Che2018.Recurrent,Chen2024.ContiFormer,Yalavarthi2024.GraFITi}.
In this work we deal with probabilistic forecasting of irregular time series. Models such as NeuralFlows~\citep{Bilos2021.Neural}, GRU-ODE~\citep{DeBrouwer2019.GRUODEBayes}, and CRU~\citep{Schirmer2022.Modeling} predict only the marginal distribution for a single time stamp.
Additionally, interpolation models like HetVAE~\citep{Shukla2022.Heteroscedastic} and Tripletformer~\citep{Yalavarthi2023.Tripletformer} can also be applied for probabilistic forecasting. However, they also produce only marginal distributions.
All the above models assume underlying distribution is Gaussian which is not the case for lots of real-world datasets.
On the other hand, Gaussian Process Regression (GPR; \citealp{Durichen2015.Multitask}), and
ProFITi~\citep{Yalavarthi2024.Probabilistica} can predict proper joint distributions.
ProFITi is not marginalization consistent because of non-separable encoder and probabilistic component.

There have been works on models for tractable and consistent marginals for fixed number of variables such as tabular data.
Probabilistic Circuits~\citep{Choi2020.Probabilistica} create a sum-prod network
on the marginal distributions in such a way that marginals are tractable and consistent.
Later, \citet{Sidheekh2023.Probabilistic} added univariate normalizing flows to the leaf nodes of the circuit for better expressivity.
However, it is not trivial to extend such circuits to deal with sequential data of variable size.
Gaussian Mixture Models (GMMs)~\citep{Duda1974PatternCA} are often used only for unconditional
density estimation, but can be extended to conditional density estimation.
They can provide tractable and consistent marginal distributions. However, GMMs are not expressive enough and often require a very large number of components to approximate even simple distributions, see Figure~\ref{fig:circles_mnf}.
Note that normalizing flow models such as~\citet{Dinh2017.Density,Papamakarios2017.Masked,Papamakarios2021.Normalizing} neither provide tractable marginals nor are applicable to varying number of variables.

Existing works have explored mixtures of normalizing flows for fixed-length sequences. For example, \citet{Pires2020.Variationala} and \citet{Ciobanu2021.Mixtures} used flows with affine coupling or masked autoregressive transformations for density estimation, while \citet{Postels2021.Go} applied them to reconstruction tasks. However, these models cannot handle dynamic sequence lengths, and their marginals are intractable. Furthermore, there has been work non-Gaussian Gaussian Processes that use Normalizing Flows on top of Gaussian processes for few shot learning~\citep{Sendera2021.Nongaussian} which is only capable of predicting a single variable/column, whereas our model is capable of predicting for multiple variables/columns, even under the presence of missing values.

%% file: contents/constructing.tex
\begin{figure*}
\centering
\begin{subfigure}[c]{0.15\textwidth}
	\centering
	\captionsetup{justification=centering}
	\includegraphics[width=\linewidth]{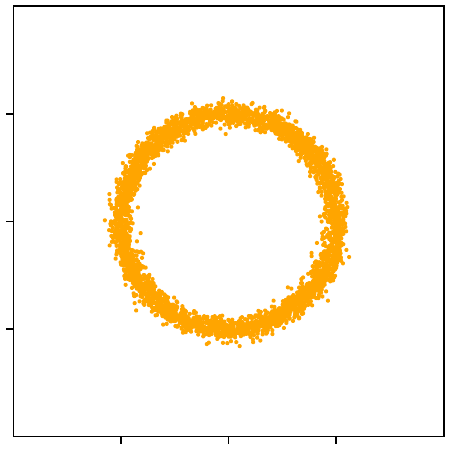}
	\subcaption*{ground truth}
\end{subfigure}
\begin{subfigure}[c]{0.55\textwidth}
\centering
\begin{subfigure}{0.2\linewidth}
	\centering
	\captionsetup{justification=centering}
	\subcaption*{$\MyModel$ (1)}
	\includegraphics[width=\linewidth]{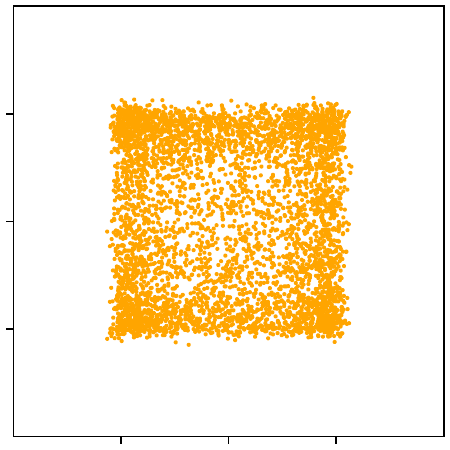}
\end{subfigure}
\begin{subfigure}{0.2\linewidth}
	\centering
	\captionsetup{justification=centering}
	\subcaption*{$\MyModel$ (2)}
	\includegraphics[width=\linewidth]{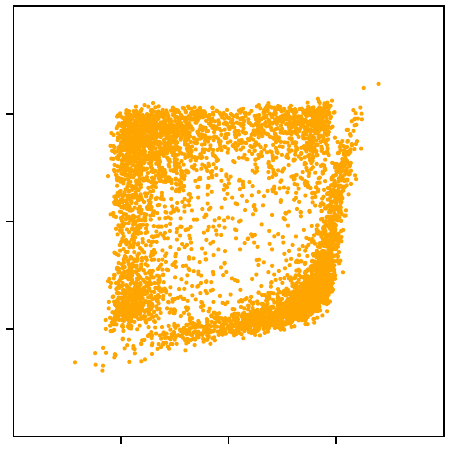}
\end{subfigure}
\begin{subfigure}{0.2\linewidth}
	\centering
	\captionsetup{justification=centering}
	\subcaption*{$\MyModel$ (3)}
	\includegraphics[width=\linewidth]{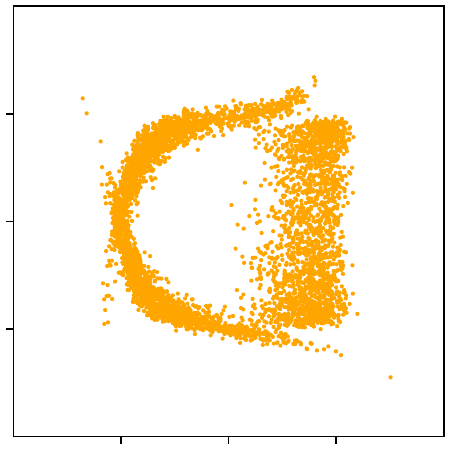}
\end{subfigure}
\begin{subfigure}{0.2\linewidth}
	\centering
	\captionsetup{justification=centering}
	\subcaption*{$\MyModel$ (4)}
	\includegraphics[width=\linewidth]{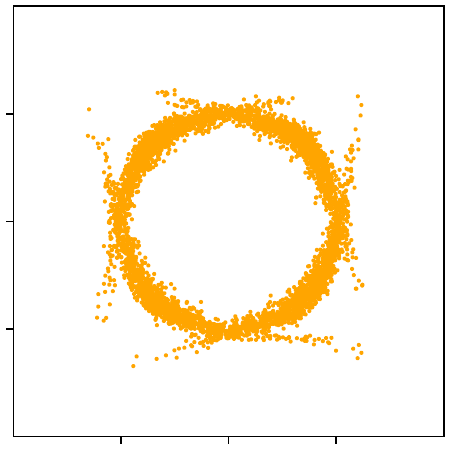}
\end{subfigure}
%
%
%
\begin{subfigure}{0.2\linewidth}
	\centering
	\captionsetup{justification=centering}
	\includegraphics[width=\linewidth]{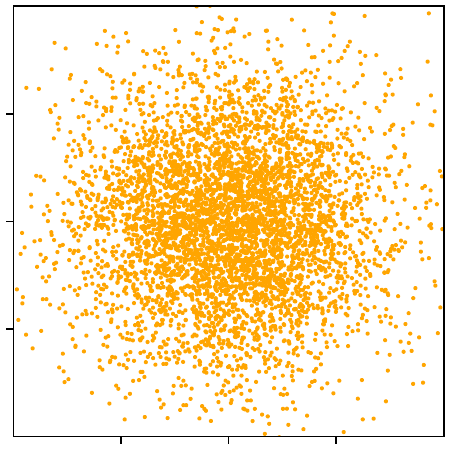}
	\subcaption*{GMM (1)}
\end{subfigure}
\begin{subfigure}{0.2\linewidth}
	\centering
	\captionsetup{justification=centering}
	\includegraphics[width=\linewidth]{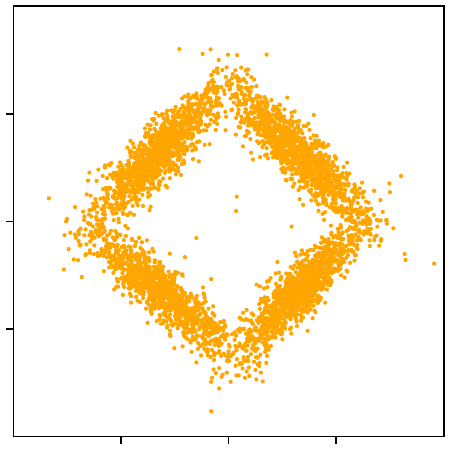}
	\subcaption*{GMM (5)}
\end{subfigure}
\begin{subfigure}{0.2\linewidth}
	\centering
	\captionsetup{justification=centering}
	\includegraphics[width=\linewidth]{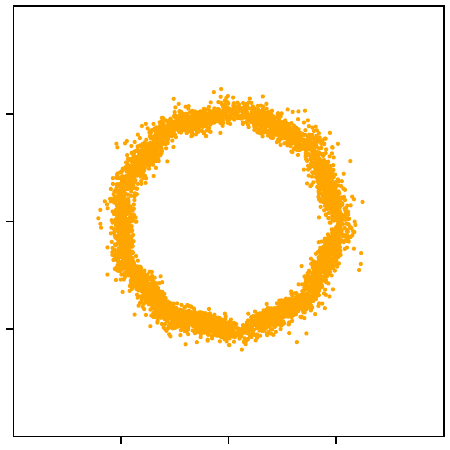}
	\subcaption*{GMM (10)}
\end{subfigure}
\begin{subfigure}{0.2\linewidth}
	\centering
	\captionsetup{justification=centering}
	\includegraphics[width=\linewidth]{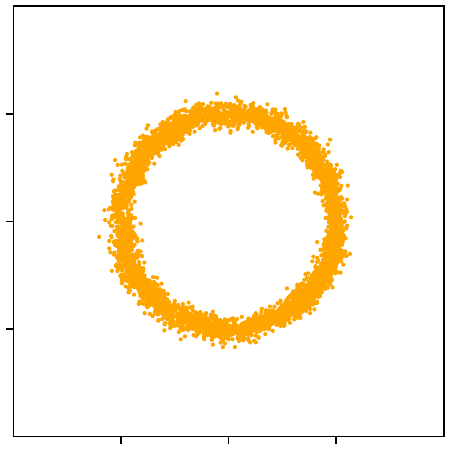}
	\subcaption*{GMM (15)}
\end{subfigure}
\end{subfigure}
\caption{(Top) Importance of multiple flow components: $\MyModel(1)$ cannot represent the correct distribution, but $\MyModel(4)$ can. (Bottom) Limitation of Gaussian Mixture Models: GMM needs 15 components to match the distribution of $\MyModel(4)$.}%
\label{fig:circles_mnf}
\end{figure*}

\section{Constructing Marginalization Consistent Conditional Distributions}\label{sec:mc_layers}

Our goal is to build a model for the conditional joint distribution
${p(y_1,\ldots,y_K \mid Q_1, \ldots, Q_k, X)}$, as in Equation~\eqref{eq:forecasting_model}. Since the model should satisfy~\ref{req:mar_consistency}, it follows that the marginal distribution of $y_k$ must only depend on $Q_k$ and $X$.

\paragraph{Separably Parametrized Gaussians.} The arguably most simple model for a permutation invariant conditional distribution for variably many variables is the family of multivariate Normal distributions ${\mathcal{N}(y \mid \mu(x), \Sigma(x))}$, whose conditional mean function $\mu(x)$ and conditional covariance function $\Sigma(x)$ are separable, i.e.:%
\begin{align}\label{eq:separable_gaussian}
	\mu_k &= \tilde{\mu}(Q_k, X)
&	\Sigma_{k,\ell} &= \widetilde{\Sigma}(Q_k, Q_{\ell}, X)
\end{align}
%
with mean function $\tilde{\mu}$ and a covariance function ${\widetilde{\Sigma}}$, a setup very well known from Gaussian processes.
Such a separably parametrized multivariate Gaussian is marginalization consistent by design, as marginalizing a Normal distribution boils down to relevant rows and columns of the covariance matrix and the corresponding elements of the mean vector.
However, Gaussian Processes form a restrictive class of models, as any joint distribution of variables is Gaussian.
To model more complex distributions, normalizing flows are a popular choice~\citep{rezende_variational_inference_normalizing_2015}.

\paragraph{Separable Normalizing Flows.}


Normalizing flows model distributions by transforming a source distribution $p_Z$ on $\R^K$
using an invertible transformation $f\colon \R^K\to \R^K$.
Then the target distribution, the distribution of the image of $f$, can
be described by the transformation theorem for densities
\begin{align}\label{eq:change_of_variable}
	p_Y(y) := p_Z(f^{-1}(y; \theta))
	\cdot \left|\det\left(\pdv{f^{-1}(y;\theta)}{y}\right)\right|%
\end{align}
Existing approaches to normalizing flows use very simple source distributions,
typically a multivariate standard normal ${p_Z(z):={\cal N}(z \mid 0, \mathbb{I})}$,
and model interactions between variables by means of the transformation~\citep{rezende_variational_inference_normalizing_2015,Papamakarios2021.Normalizing}.
Current approaches for conditional normalizing flows for a variadic number of variables
followed the same approach and tackled the problem by engineering expressive
transformations between vectors of same size,
for any size~\citep{Liu2019.Graph, 
	      Bilos2021.Normalizing,  
		  Yalavarthi2024.Probabilistica}.
For example, ProFITi uses an invertible attention mechanism.
All these models in general will not have a guarantee for marginalization
consistency. To the best of our knowledge, there is no simple condition on
the transform that would provide such a guarantee.

We therefore propose a drastic change, reversing the standard approach
for normalizing flows: to combine
  (i) simple, separable transforms with
  (ii) a richer source distribution, namely a Gaussian Process with full covariance matrix.
This way interactions between variables can be represented by the covariance of the source distribution but not by transformation of source distribution.

\begin{lemma}\label{lemma:separable_flow}
A conditional flow model over $\mathcal{\R}^K$ or $\Seq(\R)$ is \emph{separable}, if it is expressed in the form
\begin{align}\label{eq:separable_flow}
	f(z \mid Q, X) = (\phi(z_1\mid Q_1, X), \ldots, \phi(z_K\mid Q_K, X))
\end{align}
for some univariate function $\phi\colon \R\times\mathcal{Q}\times\Seq(\mathcal{X})\to \R$, that is invertible in the first argument. Any model that consists of such a separable flow transformation, combined with a marginalization consistent model for the source distribution, is itself marginalization consistent. (Proof: Appendix~\ref{proof:separable_flow})
\end{lemma}

\paragraph{Conditional Mixtures of Flows.}
When using separably parametrized Gaussians as source distributions in Lemma~\ref{lemma:separable_flow}, and expressive univariate transformations, we can model any kind of marginal as well as rich interactions between variables.
However, the model is still restricted in its expressiveness, allowing for variable-wise separable transformations of a unimodal (Gaussian) distribution only.
We therefore resort to the most simple way to further increase the expressiveness of the model: we combine several of such separable flows into a mixture.
Figure~\ref{fig:circles_mnf} shows that even just a few components can lead to a much more expressive model, in particular comparable to a simple GMM without flow transformations (more details are provided in Appendix~\ref{sec:exp_density_est}.
\begin{lemma}\label{lemma:mixture}
Given probabilistic models $(\hat{p}_d)_{d=1:D}$ that satisfy~\ref{req:joint_prediction}-\ref{req:mar_consistency}, then a mixture model
\begin{align}\label{eq:mixture}
  \hat{p}(y \mid Q, X) = \sum_{d=1}^D w_d(X) \, \hat{p}_d(y \mid Q, X)
\end{align}
with permutation invariant weight function ${w\colon \Seq(\mathcal{X})\to \Delta^D}$, were $\Delta^D$ denotes probability simplex in $D$ variables: ${\Delta^D := \{w\in \R^D\mid w_d\ge0, \sum_d w_d = 1\}}$, also satisfies~\ref{req:joint_prediction}-\ref{req:mar_consistency}. (Proof: Appendix~\ref{proof:mixture})
\end{lemma}

%% file: contents/prop_model.tex
\section{Mixtures of Separable Flows (\MyModel)}\label{sec:mix_flows}

\begin{figure*}
\centering
\resizebox{0.8\linewidth}{!}{
\import{fig/tikzpictures}{model_arch}
}
\caption{Illustration of $\MyModel$.
	$D$-many flows (fixed). $K$-many variables (variable).
	Encoder (enc) takes $\X, \Q$ (observed series and query timepoint-channel ids.) as input, and outputs
	an embedding $\hs$ (depends on both $\X$, and $\Q$) and $w$ (depends on $\X$ only). $\mu, \Sigma$ of $p_{Z_d}$ are
	parametrized by $\hs_d$.  Flow transformation of $p_{Z_d}$: parametrized
	by $\hs_d$. Transformation layer: $K$-many univariate
	transformations $\phi$ that transforms $z_k$ of $z\sim p_{Z_d}(z\mid \hs_D)$ to $y_k$ of $y\sim p_d^\flow(y\mid \hs_d)$.
}\label{fig:propmodel}
\end{figure*}

Based on the constructions from the last section, we propose to build
a marginalization consistent model for forecasting irregular time series in
four components (see Figure~\ref{fig:propmodel}):
\begin{enumerate}
\item A separable encoder, consisting of
	\begin{enumerate}
	\item A shared encoding ${\hs^\obs := \text{enc}^\obs(X; \theta^\obs)}$ of the observations,
		used for all queries.
	\item $D$-many encodings ${\hs_{d,k}:=\text{enc}^\qry(Q_k, X; \theta_d^\qry)}$
		of each query and entire context.
	\end{enumerate}
\item $D$-many Gaussian Processes $p_{Z_d}(z \mid \mu_d, \Sigma_d)$,
	each separably parametrized according to~\eqref{eq:separable_gaussian},
	by the encoder for queries $\hs_d$.
\item $D$-many separable normalizing flows $\hat{p}_d^\flow$,
	one on top of each of the source distributions,
	whose transformations $f_d$ are also separably parametrized by the encoded queries $\hs_d$.
\item A mixture of the $D$-many normalizing flows with mixing weights
    ${w:= w(\hs^\obs)}$, depending only on the encoded observations $\hs^\obs$, but not the queries.
\end{enumerate}

\paragraph{1. Separable Encoder.}
To encode both the observations ${\X = ((t_n^\obs, c_n^\obs, v_n^\obs))_{n=1:N}}$ and queries ${\Q = ((t_k^\qry, c_k^\qry))_{k=1:K}}$, we apply a positional embedding with learnable parameters $(a_f, b_f)_{f=1:F}$ to the time component~\citep{kazemi_2019_time2vec_preprint}.
\begin{align}\label{eq:positional_embedding}
\posembed(t)_f :=
\begin{cases}
	a_{f}t + b_f & \text{if \(f = 1\)}
	\\ \sin(a_{f}t + b_f) & \text{else}
\end{cases}
\end{align}
And one-hot encodings for the channel component. The value is simply passed through.
\begin{subequations}\label{eq:encodings}
\begin{alignat}{2}
	\xs &:= [ \posembed(t_n^\obs), \onehot(c_n^\obs), v_n^\obs]_{n=1:N}
\\	\qs &:= [ \posembed(t_k^\qry), \onehot(c_k^\qry) ]_{k=1:K}
\end{alignat}
\end{subequations}
The observations $\xs \in \R^{N\times(F+C+1)}$ are further encoded via self-attention and the queries $\qs \in \R^{K\times(F+C)}$ via
cross-attention w.r.t.\ the encoded observations:
\begin{subequations}\label{eq:latent_embedding}
\begin{alignat}{2}
	\hs^\obs &:= \mha(\xs, \xs, \xs; \theta^\obs)				&&\quad(\in \R^{N\times M})
\\	\widetilde{\hs} &:= \mha(\qs, \hs^\obs, \hs^\obs; \theta^\qry)	&&\quad(\in \R^{K\times D\cdot M})
\\ \hs &:= \text{reshape}(\widetilde{\hs}) &&\quad(\in \R^{D\times K\times M})
\end{alignat}
\end{subequations}
where $\mha$ denotes multihead attention. For the encoding of the queries we use an encoding dimension $D\cdot M$ and reshape each $\hs_k$ into $D$ encodings $\hs_{d,k}$ of dimension $M$.

\paragraph{2. $D$ separably parametrized Gaussian source distributions
  $p_{Z_d}(z \mid \mu_d, \Sigma_d)$.}
We model means and covariances simply by a linear and a quadratic function
in the encoded queries $\hs_d$:
\begin{subequations}\label{eq:mu_sigma_func}
\begin{align}
	\label{eq:mu_func}
	\mu(\hs_d) &= \hs_d \theta^\mean
	\implies \mu(\hs_d)_k =  \hs_{d,k}\theta^\mean
\\	\notag
	\Sigma(\hs_d) &= \bI_K + \frac{(\hs_d\theta^\cov)(\hs_d\theta^\cov)^T}{\sqrt{M'}}
	\\\label{eq:sigma_func} \implies& \Sigma(\hs_d)_{k,l}
	= \delta_{kl} + \frac{(\hs_{d,k}\theta^\cov)(\hs_{d,l}\theta^\cov)^T}{\sqrt{M'}}
\end{align}
\end{subequations}
where \(\theta^\mean \in \mathbb{R}^{M \times 1}\) and \(\theta^\cov \in \mathbb{R}^{M \times M'}\) are trainable weights shared across all \(D\) mixture components. \(\bI_K\) is the \(K \times K\) identity matrix, and \(\delta_{kl} = 1\) if \(k = l\), else \(0\), denotes the Kronecker delta.
To ensure stable learning in~\eqref{eq:sigma_func}, we scale the inner product by \(\sqrt{M'}\), following~\citep{Vaswani2017.Attention}. Since \(\Sigma(\hs_d)\) is the sum of a positive semi-definite and a positive definite matrix, it remains positive definite. Notably, \(\hs_d\) encodes both context \(X\) and queries \(Q\), assuming their roles in~\eqref{eq:separable_gaussian}.
%
%

\paragraph{3. $D$ separable normalizing flows $\hat{p}_d^\flow$.}
To achieve separable invertible transformations, any univariate bijective functions
can be applied on each variable separately.
Spline based functions attracted interest due to their expressive and generalization capabilities~\citep{Durkan2019.Neural,Dolatabadi2020.Invertible}.
We employ computationally efficient Linear Rational Spline (LRS)
transformations~\citep{Dolatabadi2020.Invertible}.
For a conditional LRS $\phi(z_k; \hs_{d,k}, \theta^\flow)$, the function parameters such as width and height of each bin, the derivatives at the knots, and $\lambda$ are computed from the conditioning input $\hs_{d,k}$ and some model parameters $\theta^\flow$.
$\theta^\flow$ helps to project $\hs_{d,k}$ to the function parameters, and is common to all the variables $z_{1:K}$ so that the transformation $\phi$ can be applied for
varying number of variables $K$. Note that we also share the same $\theta^\flow$ across all the $D$-many mixture components as well.
For details, see Appendix~\ref{appendix:lrs}.


\paragraph{4. Mixture Model.}
We model the mixture weights via cross attention,
using trainable parameters $\beta\in \R^{D\times M}$ as attention queries,
and a softmax to ensure the weights to sum to $1$:
\begin{align}\label{eq:mixture_weights}
	w := \softmax(\mha(\beta, \hs^\obs, \hs^\obs; \theta^\mix))
\end{align}
\begin{theorem}\label{thm:main_theorem}
Our model, $\MyModel$, satisfies~\ref{req:joint_prediction}-\ref{req:mar_consistency} and hence realizes a stochastic process via Kolmogorov's Extension Theorem (see Theorem~\ref{thm:kolmogorov_extension}). Proof. See Appendix~\ref{proof:marginal_consistency}.
\end{theorem}

\paragraph{Computational Complexities.}

The $D$-separable flows are computationally efficient: since they are separable, their Jacobian matrix is diagonal and computing determinant requires $\mathcal{O}(K)$ operations. The main computational cost lies in evaluating $\Sigma_d^{-1}$ and $\det \Sigma_d^{-1}$ for the base distribution, which typically requires $\bigO(K^3)$ operations.
However, for large $K$, our low-rank modification $\Sigma_d = \mathbb{I}_K + UU^T$ (see~\eqref{eq:sigma_func}) reduces their computation to $\bigO(M'^2 K)$ using the Woodbury and Weinstein–Aronszajn identities. This approach scales well for large values of $K \gg M'$, as $M'$ is independent of $K$.

\paragraph{Training.}
Given a batch $\mathcal{B}$ of training instances ($\Q, \X, y$), we minimize the normalized joint negative log-likelihood (njNLL)~\citep{Yalavarthi2024.Probabilistica}:
\begin{align}\label{eq:loss}
\mathcal{L}^\text{njNLL}(\theta) = \frac{1}{|\mathcal{B}|} \sum_{(Q, X, y)\in \mathcal{B}}
	-\frac{1}{|y|} \log\hat{p}(y\mid Q, X)%
\end{align}
where $\theta := (\theta^\obs, \theta^\qry, \theta^\mix, \theta^\mean,
\theta^\cov, \theta^\flow)$.
njNLL generalizes NLL to dynamic size variables.

%% file: contents/fig/tikzpictures/model_arch.tex
\begin{tikzpicture}
	\node at (0,0) (N_1) {\tikzlbl{$p_{Z_1}(z\mid \hs_1)$ \\ \\ $\mathcal{N}(\mu(\hs_1), \Sigma(\hs_1))$}}; 

	\node[right of=N_1, node distance=4cm] (N_D) {\tikzlbl{$p_{Z_d}(z\mid \hs_d)$ \\ \\$\mathcal{N}(\mu(\hs_d), \Sigma(\hs_d))$}}; 

	\node[draw, fill=red!10, above of=N_1, node distance=1.4cm] (flow_1) {$f_1$}; 

	\node[draw, fill=red!10, above of=N_D, node distance=1.4cm] (flow_D) {$f_d$}; 

	\node[above of=flow_1, node distance=1cm] (y_1) {$p_{Y_1}(y\mid \hs_1)$}; 
	\node[above of=flow_D, node distance=1cm] (y_D) {$p_{Y_d}(y\mid \hs_d)$}; 

	\node[above =of $(y_1.north)!0.5!(y_D.north)$, node distance=2cm, inner sep=0, label={$p_Y(y \mid \Q, \X)$}] (prod) {\large $\bigoplus$}; 

	\node[draw=none, left of=prod, node distance=3.8cm, minimum height=0.5cm] (xobs) {$\X$}; 
	\node[draw=none, left of=xobs, node distance=1.1cm, minimum height=0.5cm] (xqu) {$\Q$}; 
	\node[draw, fill=green!30, below =of $(xobs.north)!0.5!(xqu.north)$, node distance=1cm] (encoder) {$\enc$}; 
	\node[below of=encoder, node distance=1cm] (encoder_out) {\tikzlbl{$\hs \in \R^{D\times K\times M}$ \\ $w \in \Delta^D$ }}; 
	\node[draw, fill=red!10, right of=flow_D, node distance=3cm] (spline_1) {$\phi$}; 
	\node[draw, fill=red!10, right of=spline_1, node distance=1.8cm] (spline_K) {$\phi$}; 
	\node[below of=spline_1, node distance=1cm] (z_D1) {$z_1 \in \R$};
	\node[below of=spline_K, node distance=1cm] (z_DK) {$z_K \in \R$};
	\node[above of=spline_1, node distance=1cm] (y_D1) {$y_1 \in \R$};
	\node[above of=spline_K, node distance=1cm] (y_DK) {$y_K \in \R$};
	\node[above =of $(spline_1.north)!0.5!(spline_K.north)$, node distance=1cm] (y_D_sample) {$y \sim p_{Y_d}(y\mid \hs_d)$};
	\node[below =of $(spline_1.south)!0.5!(spline_K.south)$, node distance=0.1cm] (y_D_sample) {$z \sim p_{Z_d}(z\mid \hs_d)$};

	\draw[<->, red] (N_1) -- (flow_1);
	\draw[<->, red] (N_D) -- (flow_D);
	\draw[<->, red] (flow_1) -- (y_1);
	\draw[<->, red] (flow_D) -- (y_D);
	\draw[->, blue] (y_1.45) -- node (w_1) [label=$w_1$]{}  (prod);
	\draw[->, blue] (y_D) -- node (w_D) [label=$w_d$]{} (prod);

	\draw[->] (xobs) -- (encoder); 
	\draw[->] (xqu) -- (encoder); 
	\draw[->] (encoder) -- (encoder_out); 

	\draw[->, brown] ([xshift=-0.7cm]flow_1.west) -- node [yshift=2mm]{\color{black}$h_1$} (flow_1.west);

	\draw[->, brown] ([xshift=0.7cm]flow_D.0) -- node  [yshift=0.2cm]{\color{black}$\hs_d$} (flow_D.0);

	\draw[dotted] (w_1) -- (w_D);
	\draw[dotted] (flow_1) -- node [yshift=2mm]{\color{teal}$D$-many flows} (flow_D);
	\draw[dotted] (N_1) -- (N_D);

	\draw[<->, red] (z_D1) -- (spline_1);
	\draw[<->, red] (z_DK) -- (spline_K);
	\draw[<->, red] (spline_1) -- (y_D1);
	\draw[<->, red] (spline_K) -- (y_DK);
	\draw[->, brown] ([xshift=-0.9cm]spline_1.west) -- node [yshift=2mm]{\color{black}$\hs_{d,1}$} (spline_1.west);
	\draw[->, brown] ([xshift=0.9cm]spline_K.east) -- node [yshift=2mm]{\color{black}$\hs_{d,K}$} (spline_K.east);

	\draw[dotted] (spline_1) -- node (spline_center) {} (spline_K);
	\node[draw, rounded corners, dash dot, minimum width =4cm, minimum height=3.9cm] at (spline_center) (spline_box){}; 
	\draw[->, dash dot] (flow_D.north east) -- (spline_box.140);
	\draw[->, dash dot] (flow_D.south east) -- (spline_box.220);
\end{tikzpicture}


%% file: contents/experiments.tex
\section{Experiments}\label{sec:exp}

\begin{figure*}
	\centering
	\small
	\setlength{\tabcolsep}{2pt}
		\begin{tabular}{rcccc@{\quad}rcccc}
			& ground & \multirow{1}{*}{\MyModel(1)} & \multirow{1}{*}{ProFITi} & \multirow{1}{*}{GPR} & & ground & \multirow{1}{*}{\MyModel(4)} & \multirow{1}{*}{ProFITi} & \multirow{1}{*}{GPR} \\
			& truth & (ours) &  &  & & truth & (ours) &  &
			\\
			& & \scriptsize $\MWD$ = 0.05 & \scriptsize $\MWD$ = 1.59 & \scriptsize $\MWD$ = 0.05 & & &\scriptsize $\MWD$ = 0.01 & \scriptsize $\MWD$ = 1.33 & \scriptsize $\MWD$ = 0.01
			\\
			\raisebox{0.5cm}{\rotatebox[origin=c]{90}{samples}}
			&	\includegraphics[width=0.09\textwidth]{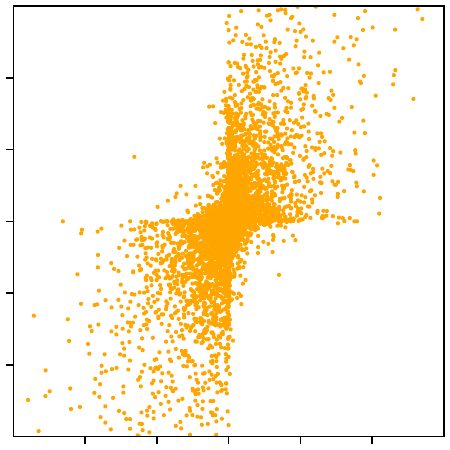}
			&	\includegraphics[width=0.09\textwidth]{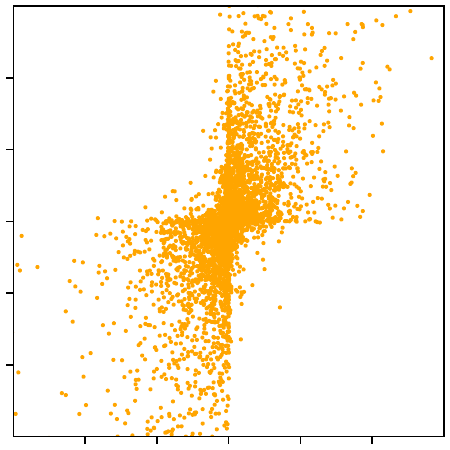}
			&	\includegraphics[width=0.09\textwidth]{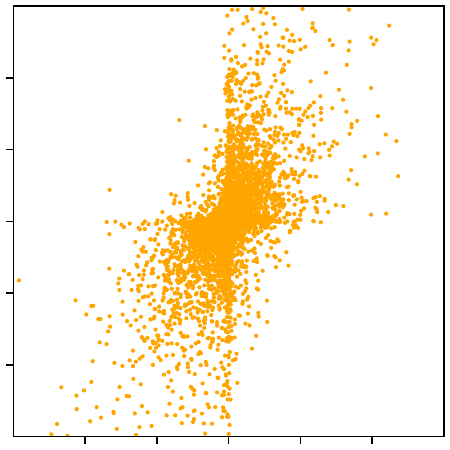}
			&	\includegraphics[width=0.09\textwidth]{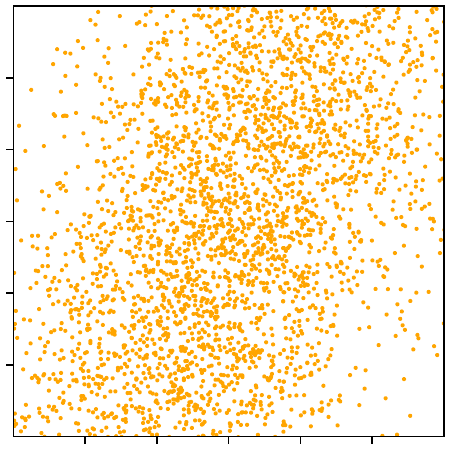}
			&	\raisebox{0.5cm}{\rotatebox[origin=c]{90}{samples}}
			&	\includegraphics[width=0.09\textwidth]{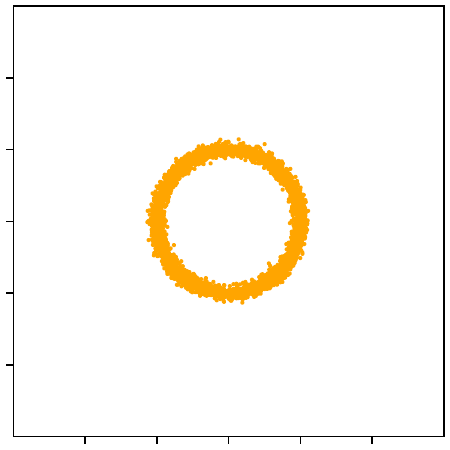}
			&	\includegraphics[width=0.09\textwidth]{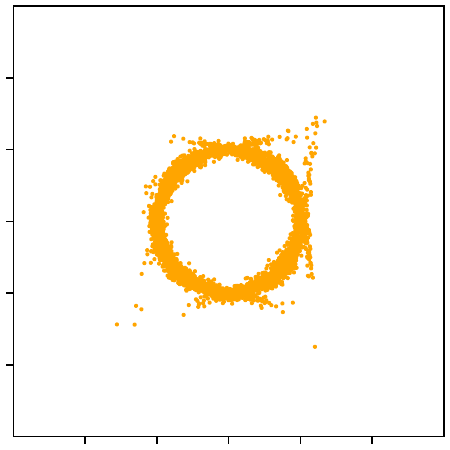}
			&	\includegraphics[width=0.09\textwidth]{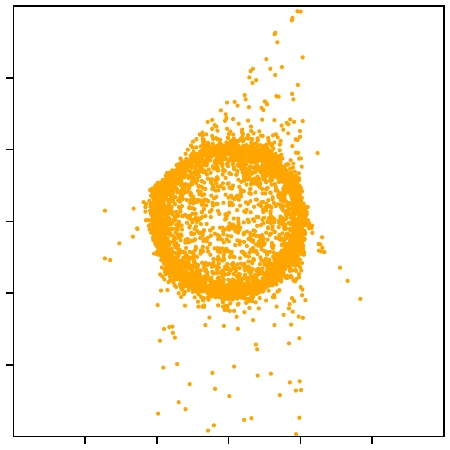}
			&	\includegraphics[width=0.09\textwidth]{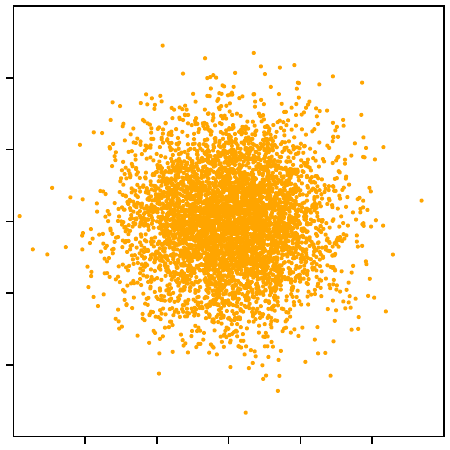}
			\\	\raisebox{0.5cm}{\rotatebox[origin=c]{90}{$p(y_1)$}}
			&	\includegraphics[width=0.09\textwidth]{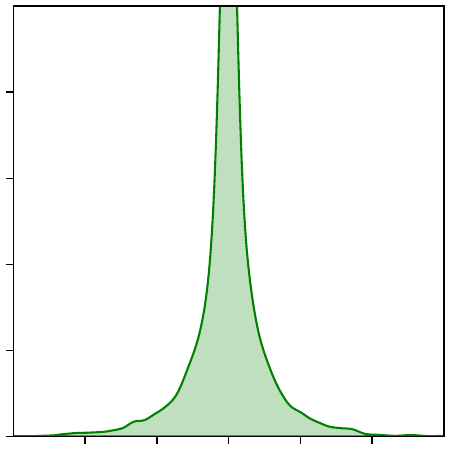}
			&	\includegraphics[width=0.09\textwidth]{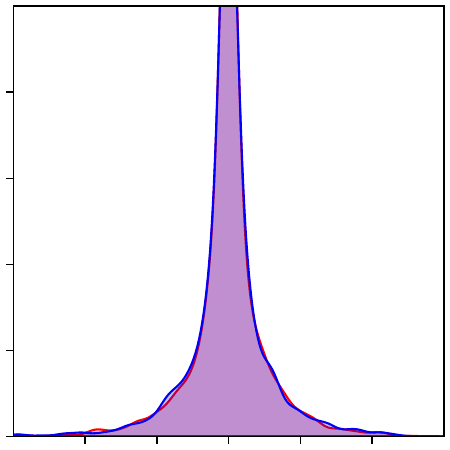}
			&	\includegraphics[width=0.09\textwidth]{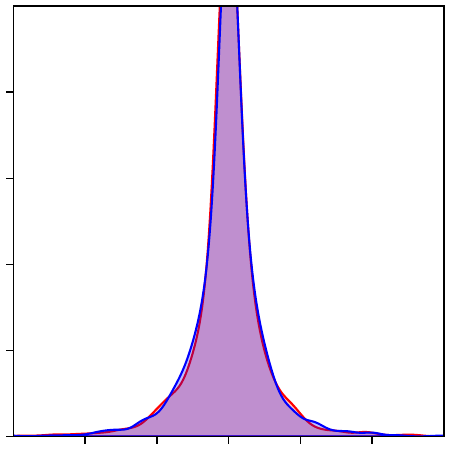}
			&	\includegraphics[width=0.09\textwidth]{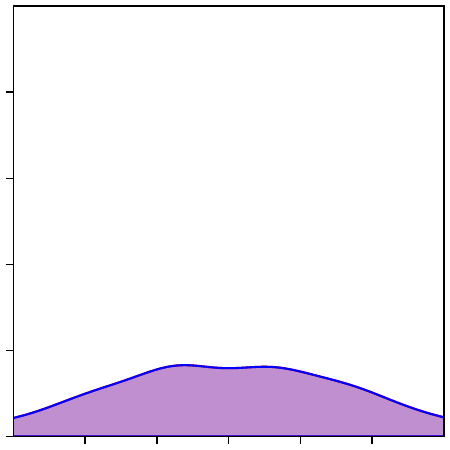}
			&	\raisebox{0.5cm}{\rotatebox[origin=c]{90}{$p(y_1)$}}
			&	\includegraphics[width=0.09\textwidth]{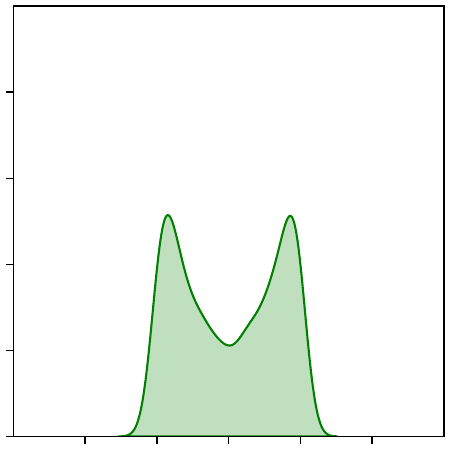}
			&	\includegraphics[width=0.09\textwidth]{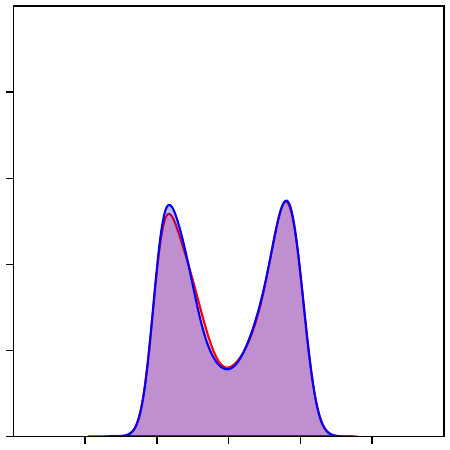}
			&	\includegraphics[width=0.09\textwidth]{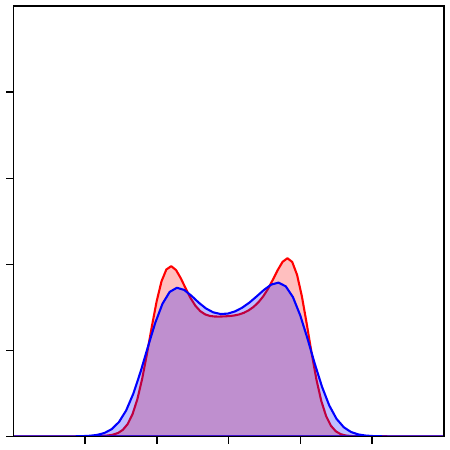}
			&	\includegraphics[width=0.09\textwidth]{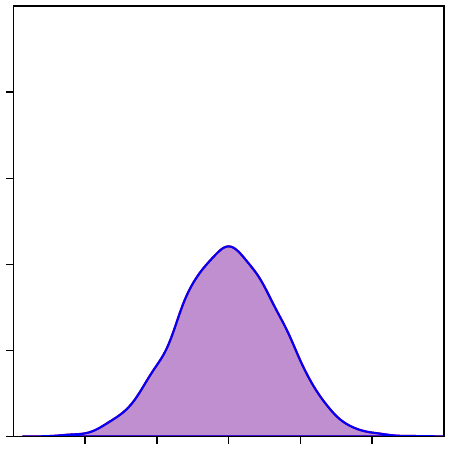}
			\\	\raisebox{0.5cm}{\rotatebox[origin=c]{90}{$p(y_2)$}}
			&	\includegraphics[width=0.09\textwidth]{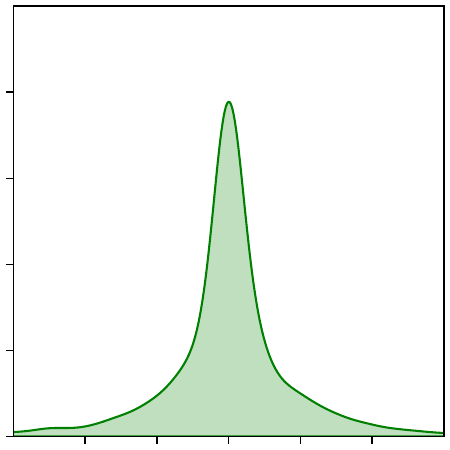}
			&	\includegraphics[width=0.09\textwidth]{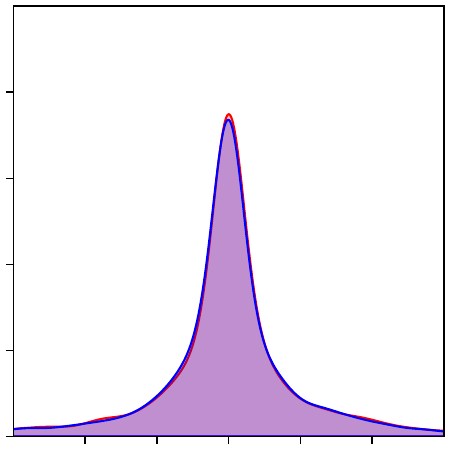}
			&	\includegraphics[width=0.09\textwidth]{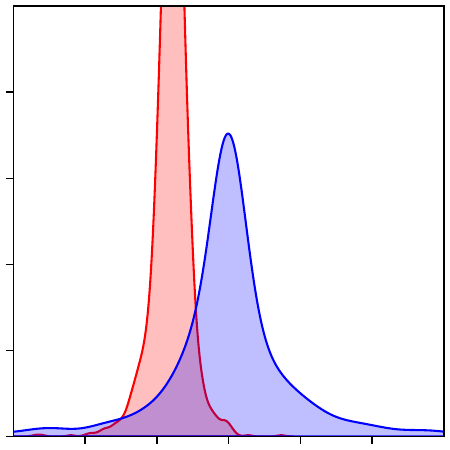}
			&	\includegraphics[width=0.09\textwidth]{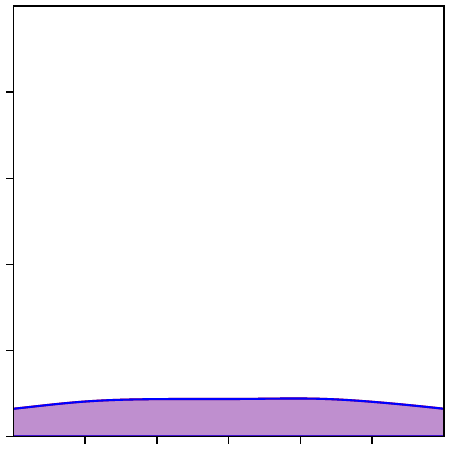}
			&	\raisebox{0.5cm}{\rotatebox[origin=c]{90}{$p(y_2)$}}
			&	\includegraphics[width=0.09\textwidth]{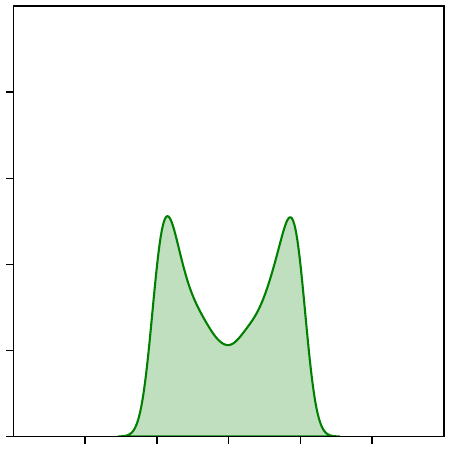}
			&	\includegraphics[width=0.09\textwidth]{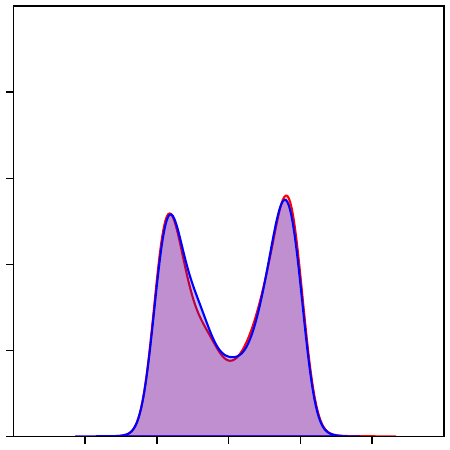}
			&	\includegraphics[width=0.09\textwidth]{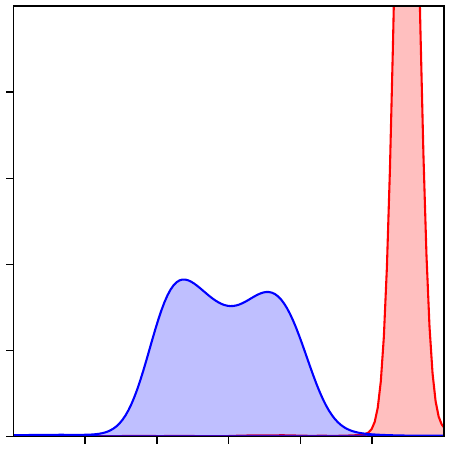}
			&	\includegraphics[width=0.09\textwidth]{fig/toy/gmix_toy_y2_both.pdf}
			\\ \multicolumn{10}{c}{\scriptsize
				\crule[teal!60!green!40]{5mm}{2mm} ground truth marginal density
				\quad\crule[red!40]{5mm}{2mm} predicted marginal density
				\quad\crule[blue!40]{5mm}{2mm} marginalized joint density
			}
		\end{tabular}
	\caption{%
		Demonstration of marginal consistency for $\MyModel$ (ours), ProFITi~\cite{Yalavarthi2024.Probabilistica}, and Gaussian Process Regression~\cite{Bonilla2007.Multitask} on two toy datasets: blast (left) and circle (right). ProFITi is inconsistent w.r.t. the marginals of the second variable $y_2$, while $\MyModel$ is consistent with the marginals of both $y_1$ and $y_2$. $\MyModel(D)$ indicates $D$ mixture components. Gaussian Process Regression (GPR) is marginalization consistent but predicts incorrect distributions.
	}%
	\label{fig:spaghetti}
\end{figure*}

\subsection{Measuring Marginalization Consistency Violation}\label{sec:mi_metric}%
We assess how well a model's predicted marginals match with those derived from its joint distribution using the \textbf{2-Wasserstein distance (WD)} or \textbf{Earth Movers Distance}.
For each variable $y_k$, we compare:
\begin{itemize}
	\item $\hat{p}(y_k \mid Q_k, X)$: the predicted marginal,
	\item $\hat{p}^\mar(y_k \mid Q_k, X)$: the marginal obtained by integrating the joint $\hat{p}(y \mid Q, X)$.
\end{itemize}
Since direct sampling from $\hat{p}^\mar$ is difficult, we sample from the joint and extract the $k$-th component.
The marginalization inconsistency is defined as the average WD across all $K$ variables:
\begin{align}
	\MI = \frac{1}{K} \sum_{k=1}^{K} \text{WD}\left( \hat{p}(y_k \mid Q_k, X), \hat{p}^\mar(y_k \mid Q_k, X) \right) \label{eq:marginal_inconsistency}
\end{align}
We use 1000 samples to compute the WD. Currently, we only use univariate marginals for computation. In principle, multivariate marginals could also be used, however they are computationally prohibitively expensive due to a lack of a closed form solution to compute the distance.

\paragraph{Toy experiment.}

We demonstrate that $\MyModel$ satisfies marginalization consistency using two synthetic bivariate distributions (Blast and Circle; see Figure~\ref{fig:spaghetti}, equations in Appendix~\ref{sec:supp_dsets}). The task is to estimate the unconditional joint distribution. $\MyModel$ accurately models both joint and marginal distributions while preserving consistency. In contrast, ProFITi captures the joint distribution well—especially for Blast—but fails on marginals due to its triangular attention mechanism, which enforces a fixed dependency order. GPR maintains consistency but lacks predictive accuracy.


\begin{table*}
\centering
\small
\caption{Comparing njNLL. Lower the better, best results in bold, second best in italics.
}\label{tab:irreg_forec}
\begin{tabular}{c@{}lrrrr}
	\toprule[1.5pt]
	&{Model}	&	\multicolumn{1}{c}{USHCN}	&	\multicolumn{1}{c}{Physionet'12}	&	\multicolumn{1}{c}{MIMIC-III}	&	\multicolumn{1}{c}{MIMIC-IV}	\\
	\midrule
	\multirow{1}{*}{inconsistent}	&ProFITi	&	\textit{-3.226 ± 0.225}	&	\textbf{-0.647 ± 0.078}	&	\textbf{-0.377 ± 0.032}	&	\textbf{-1.777 ± 0.066}\\
	\midrule
	&GRU-ODE	&	0.766 ± 0.159	&	0.501 ± 0.001	&	0.961 ± 0.064	&	0.823 ± 0.318	\\

	{consistent} &NeuralFlows	&	0.775 ± 0.152	&	0.496 ± 0.001	&	0.998 ± 0.177	&	0.689 ± 0.087	\\

	{univariate}&CRU	&	0.761 ± 0.191	&	1.057 ± 0.007	&	1.234 ± 0.076	&	\multicolumn{1}{c}{OOM}	\\
	&	Tripletformer+	&	4.632 ± 8.179	&	0.519 ± 0.112	&	1.051 ± 0.141	&	0.686 ± 0.115	\\
	\midrule
	\multirow{3}{*}{\begin{tabular}{c}consistent \\ multivariate\end{tabular}}& GPR	&	2.011 ± 1.376	&	1.367 ± 0.074	&	3.146 ± 0.359	&	2.789 ± 0.057	\\
	& GMM	&	1.050 ± 0.031	&	1.063 ± 0.002	&	1.160 ± 0.020	& 1.076 ± 0.003	\\
	\cmidrule{2-6}
	& $\MyModel$ (ours)	&	\textbf{-3.357 ± 0.176}	&	\textit{-0.491 ± 0.041}	& \textit{-0.305 ± 0.027}	&	\textit{-1.668 ± 0.097}	\\
	\bottomrule[1.5pt]
\end{tabular}
\end{table*}
\begin{table*}[t]
	\centering
	\small
	\caption{Trained for njNLL and evaluate for mNLL, lower the
		better.}%
	\label{tab:mnll}
	\begin{tabular}{c@{}lrrrr}
		\toprule
		&{Model} & \multicolumn{1}{c}{USHCN} & \multicolumn{1}{c}{Physionet'12}       & \multicolumn{1}{c}{MIMIC-III} & \multicolumn{1}{c}{MIMIC-IV} \\
		\midrule
	\multirow{1}{*}{inconsistent}	& ProFITi	&	{\textit{-3.324 ± 0.206}}		&	\textit{-0.016 ± 0.085} &	\textit{0.408 ± 0.030}	&	\textit{0.500 ± 0.322}\\
		\midrule
		& GRU-ODE        & 0.776 ± 0.172 & 0.504 ± 0.061  &  0.839 ± 0.030  & 0.876 ± 0.589 \\
		{consistent} & Neural-Flows   & 0.775 ± 0.180 & 0.492 ± 0.029	& 0.866 ± 0.097 & 0.796 ± 0.053 \\
		{univariate}&CRU            & 0.762 ± 0.180 & 0.931 ± 0.019	& 1.209 ± 0.044 & \multicolumn{1}{c}{OOM} \\
		& Tripletformer+	& 0.411 ± 7.506	&	0.524 ± 0.110	&	0.894 ± 0.083	&	0.751 ± 0.063	\\
		\midrule
		\multirow{3}{*}{\begin{tabular}{c}consistent \\ multivariate\end{tabular}} &
		GPR	&	\multicolumn{1}{c}{-}	&	\multicolumn{1}{c}{-}	&	\multicolumn{1}{c}{-}	&	\multicolumn{1}{c}{-}	\\ & GMM	&	1.042 ± 0.021	&	1.069 ± 0.002	&	1.124 ± 0.007	&	1.075 ± 0.007	\\
		\cmidrule{2-6}
		& $\MyModel$ (ours)	& {\textbf{-3.355 ± 0.156}}	& \textbf{-0.271 ± 0.028}	& \textbf{0.163 ± 0.026}	&	{\textbf{-0.634 ± 0.017}}	\\
		\bottomrule
	\end{tabular}
\end{table*}

\subsection{Main experiment}
We evaluate our model on four real-world datasets: one climate dataset (USHCN) and three medical datasets (Physionet’12, MIMIC-III, and MIMIC-IV). Following prior work~\citep{Yalavarthi2024.Probabilistica,Bilos2021.Neural}, we observe the first 36h and predict the next 3 time steps for medical datasets, and observe 3 years and predict 3 time steps for USHCN. Both the number of observations ($N$) and queries ($K$) vary across samples (see Table~\ref{tab:dset}). We split each dataset into training, validation, and test sets using a 70:10:20 ratio.
We train $\MyModel$ using the Adam optimizer with a learning rate of 0.001 and batch size of 64. Hyperparameter search is over mixture components $D \in {1,2,5,7,10}$, attention heads $\in {1,2,4}$, and latent sizes $M, F \in {16,32,64,128}$. All models are implemented in PyTorch and trained on NVIDIA RTX 3090 and GTX 1080 Ti GPUs.

\paragraph{Baselines.}
As baseline models, we use NeuralFlows~\citep{Bilos2021.Neural}, GRU-ODE~\citep{DeBrouwer2019.GRUODEBayes}, CRU~\citep{Schirmer2022.Modeling}, GPR~\citep{Durichen2015.Multitask}, and ProFITi~\citep{Yalavarthi2024.Probabilistica}.
Our encoder is similar to Tripletformer~\citep{Yalavarthi2023.Tripletformer} that predict marginal distributions for interpolation. We used it for the forecasting and called the model Tripletformer+. NeuralFlows, GRU-ODE, CRU, and Tripletformer+ predict only marginals and are marginalization consistent, as their joint distribution is the product of marginals. GPR is also marginalization consistent. We also compare with Gaussian Mixture Model (GMM) which is $\MyModel$ without flows attached to highlight the advantage of flows in $\MyModel$.


\begin{figure}[t]
\centering
\scriptsize
\begin{tabular}{rllll}
	\multicolumn{5}{c}{\resizebox{0.8\linewidth}{!}{\import{fig/tikzpictures}{wd_labels}}}
\\	\raisebox{1cm}{\rotatebox[origin=c]{90}{njNLL $\downarrow$}}
&	\import{fig/tikzpictures}{wd_ushcn}
&	\import{fig/tikzpictures}{wd_physionet}
&	\import{fig/tikzpictures}{wd_mimicii}
&	\import{fig/tikzpictures}{wd_mimiciv}
\\
&	\multicolumn{1}{c}{$\MWD\!\downarrow$}
&	\multicolumn{1}{c}{$\MWD\!\downarrow$}
&	\multicolumn{1}{c}{$\MWD\!\downarrow$}
&	\multicolumn{1}{c}{$\MWD\!\downarrow$}
\\
&	\multicolumn{1}{c}{USHCN}
&	\multicolumn{1}{c}{Physionet'12}
&	\multicolumn{1}{c}{MIMIC-III}
&	\multicolumn{1}{c}{MIMIC-IV}
\end{tabular}
\caption{njNLL vs. $\MI$\@. $\MyModel$ is marginalization consistent within sampling error.
}\label{fig:nllvsmi}

\end{figure}



\paragraph{Results.}
To highlight the importance of Marginalization Consistency in probabilistic forecasting models we train the model for njNLL in~\eqref{eq:marginal_inconsistency} and evaluate for two metrics:
1. Normalized Joint Negative Log-Likelihood (njNLL; Table~\ref{tab:irreg_forec}) and
2. Marginal Negative Log-Likelihood (mNLL; Table~\ref{tab:mnll}).
While njNLL measures the joint density of the predicted distribution, mNLL\citep{Bilos2021.Neural,Schirmer2022.Modeling} measures the univariate marginal density.
An ideal probabilistic forecasting model should perform well on both metrics, ensuring not only accurate joint predictions but also in its marginal distributions.
$\MyModel$ outperforms all marginalization-consistent models across both metrics.
As expected, ProFITi is the best performing model for njNLL. $\MyModel$ performs comparably or slightly worse than ProFITi on njNLL. However, $\MyModel$ outperforms ProFITi  significantly on mNLL.
For USHCN, ProFITi and $\MyModel$ performs comparably, difference is within standard deviation. Figure~\ref{fig:nllvsmi} shows njNLL vs
marginal inconsistency ($\MI$).
$\MyModel$ not only achieves similar likelihoods as ProFITi,
its $\MI$ is close to $0$ where ProFITi
is up to an order of magnitude larger. Smaller values of $\MI$
for $\MyModel$ is due to sampling.
We rounded the smaller $\MI$ to 0.1.
This difference stems from ProFITi’s emphasis on learning joint distributions while overlooking marginalization consistency. When trained on large-scale joint distributions and later evaluated on a single query, ProFITi experiences notable performance degradation as observed in Figure~\ref{fig:spaghetti}. In contrast, $\MyModel$ maintains consistency, ensuring minimal loss in accuracy when queried for a single time-channel. Also, we note that the performance gains of ProFITi can be mostly attributed to its encoder.
We experimented ProFITi and $\MyModel$ keeping same encoder (ProFITi-TF), and $\MyModel$ yields better accuracy than ProFITi in both MIMIC-III and MIMIC-IV (see Table~\ref{tab:profiti-tf}).

\begin{table}[t]
	\centering
	\caption{Comparing njNLL and mNLL across datasets to verify the contribution of probabilistic component of ProFITi. `ProFITi-TF`` denotes ProFITi-Transformer using same encoder as $\MyModel$}%
	\label{tab:profiti-tf}
	\begin{tabular}{lcccc}
		\toprule
		\multirow{2}{*}{Dataset} & \multicolumn{2}{c}{njNLL} & \multicolumn{2}{c}{mNLL} \\
		\cmidrule(lr){2-3} \cmidrule(lr){4-5}
		& ProFITi-TF & $\MyModel$ & ProFITi-TF & $\MyModel$ \\ 
		\cmidrule(lr){2-3} \cmidrule(lr){4-5}
		USHCN & \textbf{\underline{-3.415±0.271}} & \underline{-3.357±0.176} & \textbf{\underline{-3.440±0.243}} & \underline{-3.355±0.156} \\
		Physionet’12 & \textbf{{-0.657±0.034}} & -0.491±0.041 & 0.017±0.042 & \textbf{-0.271±0.028} \\
		MIMIC-III & 0.516±0.111 & \textbf{-0.305±0.027} & 1.279±0.057 & \textbf{0.163±0.026} \\
		MIMIC-IV & -1.405±0.220 & \textbf{-1.668±0.097} & 0.345±0.325 & \textbf{-0.634±0.017} \\
		\bottomrule
	\end{tabular}

\end{table}

\paragraph{Consistency-Accuracy Trade-off.}
While it may seem intuitive that enforcing marginalization consistency would improve the accuracy of probabilistic forecasts, this is not always the case.
Marginalization consistency enhances the predictions' reliability by ensuring coherence across marginals.
However, achieving this often requires some modeling constraints that can slightly reduce accuracy.
In critical domains such as healthcare, where trust and interpretability are crucial, the reliability afforded by consistent models is often more valuable than marginal accuracy gains.


%% file: contents/fig/tikzpictures/wd_labels.tex
\begin{tikzpicture}
	\node[mark size=0.15cm, inner sep=0pt] (MyModelmark) at (0, 0) {\color{red!70} \pgfuseplotmark{diamond*}};
	\node (MyModel) at (0.8, 0) {$\MyModel$};
	\node[mark size=0.1cm, fill=red, inner sep=0pt] (profitimark) at (2,0) {\color{blue!60} \pgfuseplotmark{square*}};
	\node (profiti) at (2.8, 0) {ProFITi};
	\node[mark size=0.12cm, fill=red, inner sep=0pt] (crumark) at (4,0) {\color{green} \pgfuseplotmark{triangle*}};
	\node (cru) at (4.6, 0) {CRU};
	\node[mark size=0.12cm, fill=red, inner sep=0pt] (neuralflowmark) at (5.6,0) {\color{brown} \pgfuseplotmark{halfsquare right*}};
	\node (neuralflow) at (6.6, 0) {NerualFlows};
	\node[mark size=0.12cm, fill=red, inner sep=0pt] (neuralflowmark) at (7.8,0){\color{violet} \pgfuseplotmark{10-pointed star}};
	\node (gruode) at (8.6, 0) {GRU-ODE};
	\node[mark size=0.1cm, fill=purple, inner sep=0pt] (gmmmarker) at (10.0, 0) {\color{magenta} \pgfuseplotmark{pentagon*}};
	\node (gmm) at (10.8, 0) {GMM};
\end{tikzpicture}

%% file: contents/fig/tikzpictures/wd_ushcn.tex
\begin{tikzpicture}
	\begin{axis}[
		scale only axis,
		width=1.6cm,
		height=1.6cm,
		xmode=log,
		xmin=0.05, xmax=20,
		ymin=-4.1, ymax=1.5,
		ytick={-3,-1,1},
		xtick={0.1,1,10}]
		\node[mark size=0.15cm, fill=red, inner sep=0pt] (str) at (0.1, -3.357) {\color{red!70} \pgfuseplotmark{diamond*}};
		\node[mark size=0.1cm, fill=red, inner sep=0pt] (str) at (0.624, -3.3226) {\color{blue!60} \pgfuseplotmark{square*}};
		\node[mark size=0.15cm, fill=purple, inner sep=0pt] (str) at (0.1, 0.94) {\color{green} \pgfuseplotmark{triangle*}};
		\node[mark size=0.13cm, fill=purple, inner sep=0pt] (str) at (0.1, 0.55) {\color{brown} \pgfuseplotmark{halfsquare right*}};
		\node[mark size=0.1cm, fill=purple, inner sep=0pt] (str) at (0.1, 0.494) {\color{violet} \pgfuseplotmark{10-pointed star}};
		\node[mark size=0.1cm, fill=purple, inner sep=0pt] (str) at (0.1, 1.01) {\color{magenta} \pgfuseplotmark{pentagon*}};
	\end{axis}
\end{tikzpicture}

%% file: contents/fig/tikzpictures/wd_physionet.tex
\begin{tikzpicture}
	\begin{axis}[
		scale only axis,
		width=1.6cm,
		height=1.6cm,
		xmode=log,
		xmin=0.05, xmax=20,
		ymin=-1, ymax=1.5,
		ytick={-1,0,1},
		xtick={0.1,1,10}]
		\node[mark size=0.15cm, fill=red, inner sep=0pt] (str) at (0.1, -0.491) {\color{red!70} \pgfuseplotmark{diamond*}};
		\node[mark size=0.1cm, fill=red, inner sep=0pt] (str) at (3.9, -0.647) {\color{blue!60} \pgfuseplotmark{square*}};
		\node[mark size=0.15cm, fill=purple, inner sep=0pt] (str) at (0.1, 1.057) {\color{green} \pgfuseplotmark{triangle*}};
		\node[mark size=0.13cm, fill=purple, inner sep=0pt] (str) at (0.1, 0.496) {\color{brown} \pgfuseplotmark{halfsquare right*}};
		\node[mark size=0.1cm, fill=purple, inner sep=0pt] (str) at (0.1, 0.501) {\color{violet} \pgfuseplotmark{10-pointed star}};
		\node[mark size=0.1cm, fill=purple, inner sep=0pt] (str) at (0.1, 1.08) {\color{magenta} \pgfuseplotmark{pentagon*}};
		%

	\end{axis}
\end{tikzpicture}

%% file: contents/fig/tikzpictures/wd_mimicii.tex
\begin{tikzpicture}
	\begin{axis}[
		scale only axis,
		width=1.6cm,
		height=1.6cm,
		xmode=log,
		xmin=0.05, xmax=20,
		ymin=-0.6, ymax=1.5,
		ytick={0,1},
		xtick={0.1,1,10}]
		\node[mark size=0.15cm, fill=red, inner sep=0pt] (str) at (0.1, -0.305) {\color{red!70} \pgfuseplotmark{diamond*}};
		\node[mark size=0.1cm, fill=red, inner sep=0pt] (str) at (1.34, -0.377) {\color{blue!60} \pgfuseplotmark{square*}};
		\node[mark size=0.1cm, fill=purple, inner sep=0pt] (str) at (0.1, 1.090) {\color{green} \pgfuseplotmark{triangle*}};
		\node[mark size=0.13cm, fill=purple, inner sep=0pt] (str) at (0.1, 0.837) {\color{brown} \pgfuseplotmark{halfsquare right*}};
		\node[mark size=0.1cm, fill=purple, inner sep=0pt] (str) at (0.1, 0.835) {\color{violet} \pgfuseplotmark{10-pointed star}};
		\node[mark size=0.1cm, fill=purple, inner sep=0pt] (str) at (0.1, 1.160) {\color{magenta} \pgfuseplotmark{pentagon*}};
	\end{axis}
\end{tikzpicture}

%% file: contents/fig/tikzpictures/wd_mimiciv.tex
\begin{tikzpicture}
	\begin{axis}[
		scale only axis,
		width=1.6cm,
		height=1.6cm,
		xmode=log,
		xmin=0.05, xmax=20,
		ymin=-2.1, ymax=1.5,
		ytick={-2,-1,0,1.0},
		xtick={0.1,1,10}]
		\node[mark size=0.15cm, fill=red, inner sep=0pt] (str) at (0.09, -1.67) {\color{red!70} \pgfuseplotmark{diamond*}};
		\node[mark size=0.1cm, fill=red, inner sep=0pt] (str) at (0.84, -1.78) {\color{blue!60} \pgfuseplotmark{square*}};
		\node[mark size=0.13cm, fill=purple, inner sep=0pt] (str) at (0.1, 0.689) {\color{brown} \pgfuseplotmark{halfsquare right*}};
		\node[mark size=0.1cm, fill=purple, inner sep=0pt] (str) at (0.1, 0.823) {\color{violet} \pgfuseplotmark{10-pointed star}};
		\node[mark size=0.1cm, fill=purple, inner sep=0pt] (str) at (0.1, 1.18) {\color{magenta} \pgfuseplotmark{pentagon*}};
	\end{axis}
\end{tikzpicture}

%% file: contents/limitations.tex
\section{Limitations}\label{sec:limitations}

The primary limitation of $\MyModel$ lies in its structural constraints on both the encoder and the probabilistic component. These restrictions can lead to slight underperformance relative to ProFITi some times in modeling the joint distribution. This work represents an initial effort to address marginalization inconsistency, and we plan to enhance the model's flexibility and performance in future. Additionally, Mixture weights cannot depend on query $Q$.
It would seem intuitive to ``switch-on''/``switch-off'' certain components depending on the query time (i.e. short term vs long term forecast). However, \ref{req:joint_prediction}-\ref{req:mar_consistency} require the weights to be independent of $Q$.

%% file: contents/conclusion.tex

\section*{Conclusions}\label{sec:concl}
In this work, we propose $\MyModel$: a marginalization-consistent mixture of separable flows for probabilistic forecasting of irregular time series with missing values. We demonstrate how to parametrize its components for decomposability and marginalization consistency. Experimental results on four real-world irregularly sampled time series datasets show that $\MyModel$ performs similarly to the state-of-the-art ProFITi model on joint distributions but significantly outperforms it on marginal distributions, highlighting the benefit of marginalization consistency.

%% file: contents/supplementary.tex
\section{Theory}\label{appendix:theory}

\subsection{Proof of Lemma~\ref{lemma:separable_flow}}\label{proof:separable_flow}

\begin{proof}
Since $X$ is a common conditional to all the marginals, we can ignore it.
So, assume that $f$ is a separable transformation:
\begin{align}
	f(z \mid Q) &= \left(\phi(z_1 \mid Q_1), \ldots, \phi(z_K \mid Q_K)\right)
\end{align}
and that $\hat{p}_{Z}(z\mid Q)$ is marginalization consistent model.
Then, the predictive distribution is
\begin{align}
	\hat{p}(y\mid Q)
	= \hat{p}_{Z}(f^{-1}(y\mid Q)\mid Q) \cdot \left|\det\dv{f^{-1}(y\mid Q)}{y}\right|
\end{align}
Since $f$ is separable, it follows that the Jacobian is diagonal:
\begin{align}
	\dv{f^{-1}(y\mid Q)}{y}
	&= \dv{\left(\phi^{-1}(y_1\mid Q_1), \ldots \phi^{-1}(y_1\mid Q_1)\right)}{(y_1, \ldots, y_K)} \notag
\\	&= \diag \left(\dv{\phi^{-1}(y_1\mid Q_1)}{y_1}, \ldots, \dv{\phi^{-1}(y_K\mid Q_K)}{y_K}\right)
\end{align}
Hence, the determinant of the Jacobian is the product of the diagonal elements:
\begin{align}\label{eq:prod_diag}
	\left|\det\dv{f^{-1}(y\mid Q)}{y}\right|
	= \prod_{k=1:K} \left|\det\dv{\phi^{-1}(y_k\mid Q_k)}{y_k}\right|
	= \prod_{k=1:K} \left|\dv{\phi^{-1}(y_k\mid Q_k)}{y_k}\right|
\end{align}
Using this fact, we can integrate the joint density over $y_k$ to get the marginal density:
\begin{flalign*}
&	\int \hat{p}(y\mid Q) \dd{y_k}
	\\ = &\int \hat{p}_{Z}(f^{-1}(y\mid Q)\mid Q)
	\qquad \cdot \left|\det\dv{f^{-1}(y\mid Q)}{y}\right| \dd{y_k}
	&&\eqcmt{~\eqref{eq:change_of_variable}}
\\	&= \int \hat{p}_{Z}(f^{-1}(y\mid Q)\mid Q)
	 \qquad\cdot \prod_{k=1:K} \left|\dv{\phi^{-1}(y_k\mid Q_k)}{y_k}\right| \dd{y_k}
	&&\eqcmt{\eqref{eq:prod_diag}}
\\	&= \left(\prod_{l\ne k}\left|\dv{\phi^{-1}(y_k\mid Q_k)}{y_k}\right|\right)
	 \qquad \cdot\int \hat{p}_{Z}(f^{-1}(y\mid Q)\mid Q) \cdot \left|\dv{\phi^{-1}(y_k\mid Q_k)}{y_k}\right| \dd{y_k}
	\span\span
\\	&= \left(\prod_{l\ne k}\left|\dv{\phi^{-1}(y_k\mid Q_k)}{y_k}\right|\right)
	 \qquad \cdot\int \hat{p}_{Z}(z\mid Q) \dd{z_k}
	&&\eqcmt{transf.-thm}
\\	&= \left(\prod_{l\ne k}\left|\dv{\phi^{-1}(y_k\mid Q_k)}{y_k}\right|\right) \hat{p}_{Z}(z_{-k}\mid Q_{-k})
	&&\eqcmt{\eqref{eq:mar_consistency}}
\\	&= \hat{p}_{Z}(z_{-k}\mid Q_{-k}) \left|\det\dv{f^{-1}(y_{-k}\mid Q_{-k})}{y_{-k}}\right|
	&&\eqcmt{\eqref{eq:prod_diag}}
\\	&= \hat{p}(y_{-k}\mid Q_{-k})
	&&\eqcmt{~\eqref{eq:change_of_variable}}
\end{flalign*}
\end{proof}

\subsection{Proof of Lemma~\ref{lemma:mixture}}\label{proof:mixture}

\begin{proof}
Consider a mixture model of the form
\begin{align}
	\hat{p}(y \mid Q, X) := \sum_{d=1}^D w_d(X) \hat{p}_d(y \mid Q, X)
\end{align}
satisfying the conditions from Lemma~\ref{lemma:mixture}, i.e.\ the component models $\hat{p}_d$ satisfy the requirements~\ref{req:joint_prediction}-\ref{req:mar_consistency} and the weight function $w\colon \Seq(\mathcal{X})\to \Delta^D$ is permutation invariant with respect to $\X$.
\begin{enumerate}
\item $\hat{p}$ satisfies~\ref{req:joint_prediction}:
	By construction of the mixture model, it has the same domain and codomain as the component models.
\item $\hat{p}$ satisfies~\ref{req:permutation_invariance}:
	Let $\pi\in S_{|\Q|}$ and $\tau\in S_{|\X|}$, then
	\begin{align*}
		\hat{p}(y\mid \Q^\pi, \X^\tau)
		&= \sum_{d=1}^D w_d(\X^\tau) \hat{p}_d(y\mid \Q^\pi, \X^\tau)
		\\	&= \sum_{d=1}^D w_d(\X) \hat{p}_d(y\mid \Q, \X)
		\\	&		\qquad\eqcmt{permutation invariance of \(w\) and \(\hat{p}_d\)}
		\\	&= \hat{p}(y\mid \Q, \X)
	\end{align*}
\item $\hat{p}$ satisfies~\ref{req:mar_consistency}:
	\begin{align*}
	\int \hat{p}(y\mid Q, X)\dd{y_{k}}
		&= \int \sum_{d=1}^D w_d(X) \hat{p}_d(y \mid Q, X) \dd{y_{k}}
	\\	&= \sum_{d=1}^D w_d(X) \int \, \hat p_d(y \mid Q, X) \dd{y_{k}}
	\\	&= \sum_{d=1}^D w_d(X) \hat p_d(y_{-k}\mid Q_{-k}, X)
	\quad \eqcmt{\(\hat{p}_d\) is marginalization consistent}
	\\	&= \hat p_Y(y_{-k}\mid Q_{-k}, X)
	\end{align*}
\end{enumerate}
\end{proof}

\subsection{Proof of Theorem~\ref{thm:main_theorem}}\label{proof:marginal_consistency}
\begin{proof}
Due to Lemma 1, it is sufficient to show that all the component models satisfy the requirements~\ref{req:joint_prediction}-\ref{req:mar_consistency}.
Since we use Gaussian Processes as the base distribution, Lemma~\ref{lemma:separable_flow} ensures that each component model is marginalization consistent, establishing~\ref{req:mar_consistency}.
Requirement~\ref{req:joint_prediction} is by construction.
Finally, permutation invariance~\ref{req:permutation_invariance} can be seen as follows:

First, note that, by Equation~\eqref{eq:latent_embedding}, it follows that if $\hs^\obs$ is permutation equivariant with respect to $X$, and $\widetilde{\hs}$ and $\hs$ are both permutation equivariant with respect to $\Q$ and permutation invariant with respect to $\X$.
Now, let $\pi\in S_{|\Q|}$ and $\tau\in S_{|\X|}$, then, for the $d$-th component model $\hat{p}_{Y_d}(y\mid Q, X)$.
In particular, the flow satisfies $f_d^{-1}(y^\pi, Q^\pi, X^\tau) = f^{-1}(y^\pi, \hs_d^\pi) = z^\pi$. Therefore:
\begin{align*}
&	\hat{p}_{Y_d}(y^\pi \mid \Q^\pi, \X^\tau) \\
	&= \hat{p}_{Z_d}(f^{-1}(y^\pi, \Q^\pi, \X^\tau)\mid \Q^\pi, \X^\tau)
	\cdot\left|\det\dv{f^{-1}(y^\pi, \Q^\pi, \X^\tau)}{y^\pi}\right|
\\	&= \mathcal{N}(f^{-1}(y^\pi, \hs_d^\pi)\mid \mu(\hs_d^\pi), \Sigma(\hs_d^\pi))
	\cdot\left|\det\dv{f^{-1}(y^\pi, \hs^\pi)}{y^\pi}\right|
	\quad\eqcmt{by remark above}
\\ &=	\mathcal{N}(z^\pi \mid \mu(\hs_d^\pi), \Sigma(\hs_d^\pi))
	\cdot\left|\det\dv{f^{-1}(y^\pi, \hs^\pi)}{y^\pi}\right|
\\	&=	\mathcal{N}(z \mid \mu, \Sigma)
	\cdot\left|\det\dv{f^{-1}(y^\pi, \hs^\pi)}{y^\pi}\right|
	\quad\eqcmt{permutation invariance of GP}
\\	&=	\mathcal{N}(z \mid \mu, \Sigma)
	\cdot\left|\det\dv{f^{-1}(y, \hs)}{y}\right|
	\quad\eqcmt{by~\eqref{eq:prod_diag}}
\\	&=	\hat{p}_{Y_d}(y \mid \Q, \X)
\end{align*}


%
\end{proof}

\subsection{Linear Rational Splines}\label{appendix:lrs}

Linear Rational Splines (LRS) are computationally
efficient spline functions~\cite{Dolatabadi2020.Invertible}.
Formally, given a set of monotonically increasing
points ${\{(u_m, v_m)\}_{m=1:M}}$ called knots,
that is ${u_m<u_{m+1}}$ and ${v_m<v_{m+1}}$, along with
their corresponding derivatives ${\{\Delta_m > 0\}_{m=1:M}}$,
then the LRS transformation $\phi(u)$ within a bin $u \in [u_m, u_{m+1}]$ is:
\begin{align}\label{eq:lrs}
	\phi(u) =
	\begin{cases}
		\frac{%
			\alpha_m v_m(\lambda_m - \tilde{u}) + \bar{\alpha}_m \bar{v}_m\tilde{u}
		}{%
			\alpha_m(\lambda_m-\tilde{u}) + \bar{\alpha}_m\tilde{u}
		} \hfill:& 0 \le \tilde{u} \le \lambda_m
		\\[1ex]
		\frac{%
			\bar{\alpha}_m \bar{v}_m(1-\tilde{u}) + \alpha_{m+1}v_{m+1}(\tilde{u} - \lambda_m)
		}{%
			\bar{\alpha}_m(1-\tilde{u}) + \alpha_{m+1}(\tilde{u} - \lambda_m)
		} \hfill:& \lambda_m \le \tilde{u} \le 1
	\end{cases} \notag \\
	\qq{where}
	\tilde{u} = \frac{u - u_m}{u_{m+1} - u_m} \in [0, 1]
\end{align}
Here, $\lambda_m\in(0,1)$ signifies the location of automatically inserted virtual knot between $u_m$ and $u_{m+1}$ with value $\bar{v}_m$. The values of $\lambda_m$, $\alpha_m$, $\bar{\alpha}_m$ and $\bar{v}_m$ are all automatically derived from the original knots and their derivatives~\cite{Dolatabadi2020.Invertible}.
For a conditional LRS $\phi(z_k; \hs_{d,k}, \theta)$,
the function parameters such as width and height of each bin,
the derivatives at the knots, and $\lambda$ are computed from the conditioning input
$\hs_{d,k}$ and some model parameters $\theta$. $\theta$ helps to project $\hs_{d,k}$
to the function parameters, and is common to all the variables $z_{1:K}$ so that
the transformation $\phi$ can be applied for varying number of variables $K$.
Additionally, we set $\theta$ common to all the components as well.
Since, each component
has separate embedding for a variable $z_k$ ($\hs_{d,k}$),
we achieve different transformations
in different components for same variable.

In summary, the conditional flow model is separable across the query size $f = f_1 \times \cdots \times f_K$ with
\begin{align}\label{eq:mnf_flow}
	f_d(y) := f(y\mid \hs_d) &= \left(\phi(y_1, \hs_{d,1}), \ldots, \phi(y_K, \hs_{d,K})\right)
\end{align}

\begin{table*}
\centering
\small
\caption{Statistics of the datasets used in our experiments.
	Sparsity means the percentage of missing observations in the time series.
	$N$ is the total number of observations and $K$ is the number of queries
	in our experiments in Section~\ref{sec:exp}.
}\label{tab:dset}
\begin{tabular}{lccccc}
	\toprule
	Name & \#Samples & \#Channels & Sparsity & N & K
	\\ \midrule
	USHCN	     &   1100 &	5	& $77.9\%$ &	$8-322$	&	$3-6$\\
	PhysioNet'12 & 12,000 &	37	& $85.7\%$ &	$3-519$	&	$1-53$\\
	MIMIC-III    & 21,000 &	96	& $94.2\%$ &	$4-709$	&	$1-85$\\
	MIMIC-IV     & 18,000 &	102	& $97.8\%$ &	$1-1382$	&	$1-79$\\
	\bottomrule
\end{tabular}
\end{table*}

\begin{table*}
	\centering
	\small
	\caption{Comparing models w.r.t.\ MSE\@. Lower the better.}\label{tab:mse}
	\begin{tabular}{lrrrr}
		\toprule
		Model & \multicolumn{1}{c}{USHCN} & \multicolumn{1}{c}{PhysioNet'12}       & \multicolumn{1}{c}{MIMIC-III} & \multicolumn{1}{c}{MIMIC-IV} \\
		\midrule
		GRU-ODE			&	0.410 ± 0.106	&	0.329 ± 0.004	&	\textbf{0.479 ± 0.044}	&	0.365 ± 0.012	\\
		Neural-Flows	&	0.424 ± 0.110	&	0.331 ± 0.006	&	\textbf{0.479 ± 0.045}	&	0.374 ± 0.017	\\
		CRU				&	\textbf{0.290 ± 0.060}	&	0.475 ± 0.015	&	0.725 ± 0.037	&	OOM	\\
		Tripletformer+	&	0.349 ± 0.131	&	\textbf{0.293 ± 0.018}	&	0.547 ± 0.068	&	0.369 ± 0.030\\
		ProFITi	&	{0.308 ± 0.061}	&	{0.305 ± 0.007}	&	0.548 ± 0.063	&	0.389 ± 0.015	\\
		\midrule
		$\MyModel$ (ours)	&	0.411 ± 0.099	&	0.307 ± 0.006	&	0.517 ± 0.057	&	\textbf{0.342 ± 0.028}	\\
	\end{tabular}
\end{table*}

\section{Datasets}\label{sec:supp_dsets}
4 real-world datasets are used in the experiments.

\paragraph{USHCN~\cite{Menne2015.United}.} This is a climate dataset consisting of $5$ climate variables such as daily temperatures, precipitation and snow
measured over 150 years at 1218 meteorological stations in the USA\@.
Following~\cite{DeBrouwer2019.GRUODEBayes,Yalavarthi2024.Probabilistica}, we selected
1114 stations and an observation window of 4 years from 1996 until 2000.

\paragraph*{PhysioNet2012~\cite{Silva2012.Predicting}.} This physiological dataset consists of the medical records of 12,000 patients who are admitted into ICU. 37 vitals are recorded for 48 hrs. Following the protocol of~\cite{Yalavarthi2024.GraFITi,Che2018.Recurrent}, dataset consists of hourly observations in each series.

\paragraph*{MIMIC-III~\cite{Johnson2016.MIMICIII}.} This is also a physiological dataset. It is a collection of readings of the vitals of the patients admitted to ICU at Beth Israeli Hospital. Dataset consists of
18,000 instances and $96$ variables are measured for $48$ hours.
Following~\cite{DeBrouwer2019.GRUODEBayes,Bilos2021.Neural,Yalavarthi2024.Probabilistica} observations
are rounded to 30 minute intervals.

\paragraph{MIMIC-IV~\cite{Johnson2021.Mark}.} The successor of the MIMIC-III dataset. Here, $102$ variables from patients admitted to ICU
at a tertiary academic medical center in Boston are measured for $48$ hours.
Following~\cite{DeBrouwer2019.GRUODEBayes,Bilos2021.Neural,Yalavarthi2024.Probabilistica} observations, are rounded to  1 minute intervals.

\paragraph{Blast distribution (toy dataset).}
Blast distribution is a bivariate distribution which is created as follows:
\begin{align*}
	z&\sim \mathcal{N}\left(
		\begin{bmatrix}0 \\ 0\end{bmatrix},
		\begin{bmatrix} 1 & 1 \\ 1 & 2 \end{bmatrix}
	\right)
\\	y &= \text{sign}(z) \odot z \odot z
\end{align*}

\paragraph{Circle (toy dataset).}
Circle is also a bi-variate distribution.
\begin{align*}
z &\sim \mathcal{N}(0, \mathbb{I}_2) \\
y &= \frac{z}{\Vert z \Vert_2} + 0.05 \cdot \mathcal{N}(0, \mathbb{I}_2)
\end{align*}

\begin{table}[t]
	\centering
	\small
	\caption{Experiment on varying observation and forecast horizons. Evaluation metric-njNLL, Lower the better}
	\begin{tabular}{lccc}
		\toprule
		&	36/12	&	24/24	&	12/36\\
		\midrule
		ProFITi		&	\textbf{-0.768±0.041}	&	\textbf{-0.355±0.243}	&	\textbf{-0.291±0.415}\\
		\midrule
		$\MyModel$ (ours)	&	{-0.315±0.016}	&	{-0.298±0.027}	&	{-0.063±0.049}\\
		\bottomrule
	\end{tabular}%
	\label{tab:var_obs_forc}
\end{table}
\begin{table}[t]
	\centering
	\small
	\caption{Experiment on varying observation and forecast horizons. Evaluation metric-mNLL, Lower the better}
	\begin{tabular}{lccc}
		\toprule
		&	36/12	&	24/24	&	12/36\\
		\midrule
		ProFITi		&	1.376±1.764	&0.705±0.179	&2.977±2.978\\
		$\MyModel$ (ours)	&	\textbf{-0.083±0.025}	& \textbf{-0.020±0.060}	&	\textbf{0.040±0.131}\\
		\bottomrule
	\end{tabular}%
	\label{tab:var_obs_forc_marg}
\end{table}
\section{Additional Experiments}\label{sec:add_exp}

\begin{table*}
	\centering
	\small
	\caption{Comparing for Energy Score. Lower the better}
	\begin{tabular}{lcccc}
		\toprule
		&	USHCN	&	PhysioNet'12	&	MIMIC-III	&	MIMIC-IV	\\
		\midrule
		NeuralFlows	&		0.661 ± 0.059	&	1.691 ± 0.001	&	1.381 ± 0.033	&	0.982 ± 0.009	\\
		ProFITi		&	\textbf{0.452 ± 0.044}	&	\textbf{0.879 ± 0.303}	&	1.606 ± 0.168	&	\textbf{0.808 ± 0.003}	\\
		\midrule
		$\MyModel$	&	0.552 ± 0.044	&	1.599 ± 0.013	&	\textbf{1.353 ± 0.033}	&	0.906 ± 0.029\\
		\bottomrule
	\end{tabular}%
	\label{tab:energy_score}
\end{table*}
\begin{table*}
	\centering
	\small
	\caption{Comparing models w.r.t.\ CRPS score on marginals. Lower the better.
	}\label{tab:crps}
	\begin{tabular}{lcccc}
		\toprule
		Model & \multicolumn{1}{c}{USHCN} & \multicolumn{1}{c}{PhysioNet'12}       & \multicolumn{1}{c}{MIMIC-III} & \multicolumn{1}{c}{MIMIC-IV} \\
		\midrule
		Neural-flows	&	0.306 ± 0.028	&	0.277 ± 0.003	&	0.308 ± 0.004	&	0.281 ± 0.004	\\
		ProFITi &	\textbf{0.182 ± 0.007}	&	0.271 ± 0.003	&	0.319 ± 0.003	&	0.279 ± 0.012	\\
		\midrule
		$\MyModel$ (ours)	&	{0.220 ± 0.019}	&	\textbf{0.260 ± 0.002}	&	\textbf{{0.296 ± 0.005}}	&	\textbf{0.245 ± 0.010}	\\
		\bottomrule
	\end{tabular}
\end{table*}

\subsection{Comparing for Point Forecasting}\label{sec:mse}

While point forecasting is an important task in time series analysis, the goal of probabilistic forecasting is fundamentally different. Probabilistic forecasting aims to capture the full predictive distribution rather than just a single-point estimate.
Nonetheless, one might intuitively expect that the best probabilistic model would also yield the most accurate point estimates. However, this is not always the case in practice, as noted in prior works~\citep{Lakshminarayanan2017.Simplea,Seitzer2021.Pitfallsa,Rasul2021.Multivariate,Yalavarthi2024.Probabilistica}.

We compare probabilistic models in terms of point prediction accuracy using Mean Squared Error (MSE), as reported in Table~\ref{tab:mse}. Our results show that no single model consistently outperforms the others across all datasets.
We believe there are two primary reasons for this phenomenon:

(1.) MSE is related to the Negative Log-Likelihood (NLL) of a Gaussian distribution with a fixed standard deviation.
Therefore, models explicitly trained by minimizing Gaussian Negative Log-Likelihood (even if they predict more than just the mean) are naturally optimized for this metric.

(2.) Probabilistic models are trained to predict the underlying data distribution, not solely the optimal point estimate (e.g., the conditional mean).
Their objective is to accurately capture the uncertainty and dependencies in the data, which involves learning the (co)variance structure.
This focus on the full distribution can sometimes lead to point estimates that are not strictly optimized for minimizing the squared error, even if the overall probabilistic forecast is superior.

Except for $\MyModel$ and ProFITi, all the other probabilistic models are designed to predict Gaussian distributions.
Between ProFITi and $\MyModel$, their performance is comparable. ProFITi outperforms $\MyModel$ on the USHCN whereas $\MyModel$ performs better in MIMIC-IV. For, Physionet'12 and MIMIC-III they have comparable performances (difference is within standard deviation).

\subsection{Experiment on varying observation and forecast horizons}

We would like to see if $\MyModel$ is scalable to long observations and forecast horizons.
For this, we performed an experiment on varying length observation and forecasting horizons on Physionet'12 dataset and compared against the published results from~\citep{Yalavarthi2024.Probabilistica} in Table~\ref{tab:var_obs_forc}. The observation and forecasting horizons are: \{(36h, 12h), (24h, 36h), (12h, 26h)\}.

Tables~\ref{tab:var_obs_forc} and~\ref{tab:var_obs_forc_marg} present the njNLL and mNLL results for ProFITi and $\MyModel$. The results follow the trends observed in Tables~\ref{tab:irreg_forec} and~\ref{tab:mnll}. ProFITi performs best when predicting joint distributions. However, its lack of marginalization consistency leads to a severe performance drop when predicting marginal distributions. In contrast, $\MyModel$ maintains stable performance from njNLL to mNLL. While it performs slightly worse than ProFITi on njNLL, it significantly outperforms ProFITi on mNLL.

\subsection{Comparing for Energy Score}

The Energy Score between the ground truth $y$ and predicted distribution $\hat{p}_Y$ is computed as:
\begin{align}
	\text{ES}(y, \hat{p}_Y) := \mathop{\E}_{y^\prime\sim \hat{p}_Y} \Vert y - y^\prime\Vert^p_2 - \frac{1}{2} \mathop{\E}_{y^\prime, y^{\prime \prime}\sim \hat{p}_Y} \Vert y^\prime - y^{\prime \prime} \Vert^p_2,
\end{align}
where $\Vert\cdot\Vert_2$ denotes the Euclidean norm and $p\in (0,2)$ is a parameter. In our evaluation, we set $p=1$. \citet{Marcotte2023.Regionsb} demonstrated that the Energy Score is not a reliable metric for evaluating multivariate distributions. Additionally, it suffers from the curse of dimensionality, as it requires $N^K$ samples, where $K$ is the number of variables and $N$ is the number of samples required to accurately estimate a univariate distribution.

However, since many regularly sampled, fully observed multivariate time series probabilistic forecasting models use the Energy Score as an evaluation metric, we examine how $\MyModel$ compares to the best-performing inconsistent multivariate probabilistic model, ProFITi, and the consistent univariate probabilistic model, NeuralFlows in Table~\ref{tab:energy_score}. Our results show that $\MyModel$ outperforms NeuralFlows across all datasets. As shown by the njNLL metric in Table~\ref{tab:irreg_forec}, ProFITi is the best-performing model, outperforming $\MyModel$ in 3 out of 4 datasets.

\subsection{Comparing for Marginals in Terms of CRPS}\label{sec:exp_mar_nll_extra}

We compare with CRPS score in Table~\ref{tab:crps}, a widely used evaluation metric in time series forecasting.
We see that $\MyModel$ outperforms all the consistent models. It performs better than ProFITi
in $3$ out of $4$ dataset. For ProFITi and $\MyModel$, we sampled 1000 instances and computed the CRPS\@.

\begin{table}[t]
	\centering
	\small
	\caption[table]{Ablation study on PhysioNet2012}\label{tab:abl}
	\begin{tabular}{lr}
		\toprule
		Model	&	\multicolumn{1}{c}{njNLL ($\downarrow$)}	\\
		\midrule
		$\MyModel$			&	-0.491 ± 0.041 \\
		$\MyModel$--$f$		&	 1.063 ± 0.002 \\
		$\MyModel$--$\cov$	&	-0.308 ± 0.024 \\
		$\MyModel$--$w$		&	-0.451 ± 0.038 \\
		$\MyModel$ (1)		&	-0.493 ± 0.029 \\
		\bottomrule
	\end{tabular}
\end{table}

\begin{table*}
	\scriptsize
	\caption{Comparing the number of parameters and run-time per epoch for results in Table~\ref{tab:irreg_forec} for GMM and $\MyModel$, and ProFITi for reference}
	\label{tab:parameters_runtime}
	\begin{tabular}{l|cc|cc|cc|cc}
		&	\multicolumn{2}{c|}{USHCN} & \multicolumn{2}{c|}{Physionet'12} & \multicolumn{2}{c|}{MIMIC-III}	& \multicolumn{2}{c}{MIMIC-IV}	\\
		&	Parameters	& Run Time	& Parameters & Run Time	& Parameters	& Run Time	& Parameters	&	Run Time	\\
		\cmidrule(lr){2-3} \cmidrule(lr){4-5} \cmidrule(lr){6-7} \cmidrule(lr){8-9}

		ProFITi	&1,093.0K	& 3.8s	&75.8K	&42.14s	&	59.7K&66.8s	&285.9K	&70.2s	\\
		GMM	&	416.0K &	0.9s&	390.9K&	5.9s&	33.0K&	18.5s&	101.1K&21.3s	\\
		$\MyModel$ (ours)	&167.6K	& 2.4s	& 134.6K	&	14.1s &112.6K	&	25.4s&398.6K	&	33.3s	\\
	\end{tabular}
\end{table*}

\subsection{Ablation study.}\label{sec:abl}

Using Physionet'12, we show the importance of different model components. As summarized in Table~\ref{tab:abl}, the performance is reduced by removing the flows ($\MyModel-f$) which is same as GMM\@. It is expected that normalizing flows are more expressive compared to
simple mixture of Gaussians. On the other hand, by using only isotropic Gaussian as the
base distribution ($\MyModel-\cov$) model performance worsened.
Similarly, parameterizing the components weights have a slight advantage
over fixing them to $1/D$ with $D$ being the number of components.
One interesting observation is even using single component ($\MyModel(1)$)
gives similar results compared to mixture of such components.
This could be because the dataset we have may not require multiple components. We
note that we have $D=1$ in our hyperparameter space, and we select the best $D$
based on validation dataset.

\subsection{Comparing the number of parameters and runtime for GMM and $\MyModel$}

Since \(\MyModel\) is built upon GMM, Table~\ref{tab:parameters_runtime} presents the number of parameters and runtime for both \(\MyModel\) and GMM. For reference, we also include ProFITi.

The results show that GMM has a relatively low number of parameters for MIMIC-III and MIMIC-IV, whereas for USHCN and PhysioNet'12, the number of parameters is significantly higher. The primary difference between GMM and \(\MyModel\) is the inclusion of flows. Given that all other factors remain the same, \(\MyModel\) is expected to have a slightly higher number of parameters than GMM due to these additional flows. Also, the parameters for the flows are shared among all the variables and the components, their number does not grow with increase in components or variables.
However, differences in the chosen hyperparameters for GMM and \(\MyModel\) lead to some discrepancies from this expectation.
Moreover, the inclusion of flows in \(\MyModel\) results in a slightly higher runtime compared to GMM.